\documentclass[10pt,journal,letterpaper,compsoc]{IEEEtran}

\usepackage{times}

\usepackage{latexsym,bm,amsmath,amssymb}
\usepackage{graphicx}
\usepackage{subfigure}
\usepackage{multirow}
\usepackage{float}
\usepackage{placeins}
\usepackage{algorithm}
\usepackage{algorithmicx}
\usepackage{algpseudocode}
\usepackage{array}
\usepackage{xspace}
\usepackage{balance}  

\makeatletter
\DeclareRobustCommand\onedot{\futurelet\@let@token\@onedot}
\def\@onedot{\ifx\@let@token.\else.\null\fi\xspace}

\def\eg{\emph{e.g}\onedot} 
\def\ie{\emph{i.e}\onedot}

\makeatother

\newcommand{\DoubleFigureWidth}{220pt}
\newcommand{\SmallDoubleFigureWidth}{220pt}

\newcommand{\DoubleFigureWidthTiny}{180pt}
\newcommand{\QuaterFigureWidth}{124pt}
\newcommand{\QuaterFigureWidthSmall}{118pt}

\begin{document}


\title{A Revisit of Hashing Algorithms for Approximate Nearest Neighbor Search}

\author{Deng Cai
	\thanks{D. Cai is with the State Key Lab of CAD\&CG, College of Computer Science,
		Zhejiang University, Hangzhou, Zhejiang, China, 310058.
		Email: \texttt{dengcai@cad.zju.edu.cn}.}
}

\IEEEcompsoctitleabstractindextext{%
	\vspace{6mm}
	\begin{abstract} 
	Approximate Nearest Neighbor Search (ANNS) is a fundamental problem in many areas of machine learning and data mining. During the past decade, numerous hashing algorithms are proposed to solve this problem. Every proposed algorithm claims to outperform Locality Sensitive Hashing (LSH) which is the most popular hashing method. However, the evaluation of these hashing papers was not thorough enough, and the claim should be re-examined. The ultimate goal of an ANNS method is returning the most accurate answers (nearest neighbors) in the shortest time. If implemented correctly, almost all the hashing methods will have their performance improved as the code length increases. However, many existing hashing papers only report the performance with the code length shorter than 128. In this paper, we carefully revisit the problem of search with a hash index and analyze the pros and cons of two popular hash index search procedures. Then we proposed a very simple but effective two-level index structure and made a thorough comparison of eleven popular hashing algorithms. Surprisingly, the random-projection-based Locality Sensitive Hashing (LSH) ranked the first, which is in contradiction to the claims in all the other ten hashing papers. Despite the extreme simplicity of random-projection-based LSH, our results show that the capability of this algorithm has been far underestimated. For the sake of reproducibility, all the codes used in the paper are released on GitHub, which can be used as a testing platform for a fair comparison between various hashing algorithms. 
	\end{abstract}

	\begin{IEEEkeywords}
		Approximate nearest neighbor search, hashing.
	\end{IEEEkeywords}
}

\maketitle

\IEEEdisplaynotcompsoctitleabstractindextext
\IEEEpeerreviewmaketitle

\section{Introduction}

Nearest neighbor search plays an important role in many applications of machine learning and data mining. Given a dataset with $N$ entries, the cost of finding the exact nearest neighbor is $O(N)$, which is very time consuming when the data set is large. So people turn to Approximate Nearest Neighbor (ANN) search in practice \cite{AryaM93Approximate, Kleinberg97Algorithms}. Hierarchical structure (tree) based methods, such as KD-tree \cite{Bentley1975Multidimensional} ,Randomized KD-tree \cite{Silpaanan2008Optimised}, K-means tree \cite{Fukunaga1975A}, are popular solutions for the ANN search problem. These methods perform very well when the dimension of the data is relatively low. However, the performance decreases dramatically as the dimension of the data increases \cite{Weber1998AQA}.

During the past decade, hashing based ANN search methods \cite{GiVLDBIM1999VLDB,WeissTF08NIPS,KulisG09ICCV} received considerable attention. These methods generate
binary codes for high dimensional data points (real vectors) and try to preserve the similarity among the original real vectors. 
A hashing algorithm generating $l$-bits code can be regarded as contains $l$ hash functions. Each function partitions the original feature space into two parts, the points in one part are coded as 1, and the points in the other part are coded as 0. When an $l$-bits code is used, the hashing algorithm ($l$ hash functions) partitions the feature space into $2^l$ parts, which can be named as hash buckets.
Thus, all the data points fall into different hash buckets (associated with different binary codes). Ideally, if neighbor vectors fall into the same bucket or the nearby buckets (measured by the Hamming distance of binary codes), the hashing based methods can efficiently retrieve the nearest neighbors of a query point. 

One of the most popular hashing algorithms is Locality Sensitive Hashing (LSH) \cite{IndykM98STOC,GiVLDBIM1999VLDB}. LSH is a name for a set of hashing algorithms, and we can specifically design different LSH algorithms for different types of data \cite{GiVLDBIM1999VLDB}. For real vectors, random-projection-based LSH \cite{Charikar2002LSH} might be the most simple and popular one. This algorithm uses random projection to partition the feature space.

LSH is naturally data independent. Many data-dependent hashing algorithms \cite{WeissTF08NIPS,KulisG09ICCV,KulisD09NIPS,WangKC10ICML,ZhangWCL10SIGIR,HeLC10KDD,GongL11CVPR,JolyB11CVPR,XuWLZLY11ICCV,LiuWKC11ICML,NorouziF11ICML,JegouDS11PAMI,LiuWJJC12CVPR,HeoLHCY12CVPR,WeissFT12,KongL12NIPS,LinJCYL13CVPR,HeWS13CVPR,GeHK013CVPR,JinHLZLCL13ICCV,XuBLCHC13IJCAI,JinLLC14TCYB,KalantidisA14CVPR,LiuMKC14NIPS} have been proposed during the past decades and all these algorithms claimed to have superior performance over the baseline algorithm LSH. However, the evaluation of all these papers are not thorough enough, and their conclusions are questionable:

\begin{itemize}
	\item How to measure the performance of the learned binary code might be the most critical problem in designing a new hashing algorithm. A straightforward answer could be using the binary code as the index to solve the ANNS problem. The ANNS performance can then be used to evaluate the performance of the binary code. However, there are two typical procedures for search with a hash index, the ``hamming ranking" approach and the ``hash bucket search'' approach. Each approach has its pros and cons. For different datasets, different hashing algorithms favour different approaches. It is not easy to fairly compare different hashing algorithms.
	\item Most of the existing hashing paper \cite{WeissTF08NIPS,KulisG09ICCV,KulisD09NIPS,WangKC10ICML,ZhangWCL10SIGIR,HeLC10KDD,XuWLZLY11ICCV,LiuWKC11ICML,JegouDS11PAMI,LiuWJJC12CVPR,WeissFT12,LinJCYL13CVPR,HeWS13CVPR,GeHK013CVPR,JinHLZLCL13ICCV,XuBLCHC13IJCAI,JinLLC14TCYB,KalantidisA14CVPR,LiuMKC14NIPS}  uses the "hamming ranking" approach to examine the performance of various hashing algorithms.  Since there are no publicly available c++ codes, all these papers use Matlab function and the {\bf accuracy - \# of located samples} curves to compare different hashing algorithms. Thus, different hashing algorithms should be compared with the same code length, and it is impossible to compare the hashing algorithms with various non-hashing based ANNS methods.
	\item Almost all the hashing papers report that the performance increases as the code length increases.  However, most of the hashing papers \cite{WeissTF08NIPS,KulisG09ICCV,KulisD09NIPS,WangKC10ICML,ZhangWCL10SIGIR,HeLC10KDD,XuWLZLY11ICCV,LiuWKC11ICML,JegouDS11PAMI,LiuWJJC12CVPR,WeissFT12,LinJCYL13CVPR,HeWS13CVPR,GeHK013CVPR,JinHLZLCL13ICCV,XuBLCHC13IJCAI,JinLLC14TCYB,KalantidisA14CVPR,LiuMKC14NIPS} only reported the performance with the code length shorter than 128. How about we further increase the code length?
	\item Another possible approach to search with a hash index is so-called ``hash bucket search'' approach. Different from the ``hamming ranking'' approach, ``hash bucket search'' requires sophisticated data structures. If the hash bucket search approach is used, {\bf\# of located samples} is no longer a good indicator of the search time, and the {\bf accuracy - \# of located samples} curves become meaningless.
	\item Besides the ``hamming ranking" and the ``hash bucket search'' approaches, can we design better hash index search procedure so that different hashing algorithms can be fairly and easily compared?
\end{itemize}

These problems have already been raised in the literature. Joly and Buisson \cite{JolyB11CVPR} pointed out that many data-dependent hashing algorithms have the improvements over LSH, but "improvements occur only for relatively small hash code sizes up to 64 or 128 bits". The figure 4 in this paper \cite{JolyB11CVPR} shows that when 1,024-bits codes were used, LSH is the best-performed hashing algorithm. However, Joly and Buisson \cite{JolyB11CVPR} failed to provide a detailed analysis. 

There is also a public available c++ program named FALCONN on the GitHub which implements LSH algorithm. Since we can record the search time of the program, the FALCONN (LSH) can be fairly compared with other non-hashing based ANNS methods. However, the results are not very encouraging\footnote{http://www.itu.dk/people/pagh/SSS/ann-benchmarks/}. There are even researchers pointing out that "LSH is so hopelessly slow and/or inaccurate"\footnote{http://www.kgraph.org/}. Is this true that "LSH is so hopelessly slow and/or inaccurate"?

All these problems motivate us to revisit the problem of applying hashing algorithms for ANNS. This study is carried on two popular million-scale ANNS datasets (SIFT1M and GIST1M)\footnote{http://corpus-texmex.irisa.fr/}, and the findings are very surprising:
\begin{itemize}
	\item Eleven popular hashing algorithms are thoroughly compared on these two datasets. Despite the fact that all the other ten data dependent hashing algorithms claimed the superiority over LSH which is data independent, LSH ranked the first among all the eleven compared algorithms on SIFT1M and ranked the second on GIST1M. 
	\item We also compare our implementation of random-projection-based LSH (RPLSH) with other five popular open source ANNS algorithms (FALCONN, flann, annoy, faiss and KGrpah). On GIST1M, RPLSH performs best among the six compared methods. On SIFT1M, RPLSH performs the second best, and if a high recall is required (higher than 99\%),  RPLSH performs the best. 
\end{itemize} 
Despite the extreme simplicity of the random-projection-based LSH, our results show that the capability of this algorithm has been far underestimated.
For the sake of reproducibility, all the codes are released on GitHub\footnote{https://github.com/ZJULearning/hashingSearch} which can be used as a testing platform for a fair comparison between various hashing algorithms.

It is worthwhile to highlight the contributions of this paper:
\begin{itemize}
	\item The goal of this paper is not introducing yet another hashing algorithm. We aim at providing analysis on how to correctly measure the performance of a hashing algorithm. Given the fact that almost all the existing hashing papers \cite{WeissTF08NIPS,KulisG09ICCV,KulisD09NIPS,WangKC10ICML,ZhangWCL10SIGIR,HeLC10KDD,XuWLZLY11ICCV,LiuWKC11ICML,JegouDS11PAMI,LiuWJJC12CVPR,WeissFT12,LinJCYL13CVPR,HeWS13CVPR,GeHK013CVPR,JinHLZLCL13ICCV,XuBLCHC13IJCAI,JinLLC14TCYB,KalantidisA14CVPR,LiuMKC14NIPS} failed to address this problem and gave misleading conclusions, we believe such analysis is important and useful.
	\item We introduce a simple yet novel two-level index scheme to search with a hash index. This approach significantly outperforms the traditional hamming ranking and the hash bucket search approaches. With this novel search approach, it becomes easy to compare various hashing algorithms fairly. 
	\item We release an open source ANNS library RPLSH which implements this two-level index scheme with random-projection-based LSH. The comparison with other five popular open source ANNS algorithms (FALCONN, flann, annoy, faiss and KGraph) demonstrate the superiority of RPLSH.  It performs the best on GIST1M and performs the second best on SIFT1M. 
	\item Recently, graph-based algorithms \cite{Ben2016Fanng, MalkovNMSLIB2016, li2016approximate, fu2017fast} have shown very promising performance on the ANNS problem. Besides KGraph, we do not compare with other more advanced graph-based algorithms. We agree that the performance of RPLSH is inferior to the performance of some graph-based algorithms (\eg, HNSW\cite{MalkovNMSLIB2016}, NSG\cite{fu2017fast}) on CPU. However, RPLSH has its own advantages: very small index size, very efficient indexing and very simple search procedure.  In some cases the graph-based algorithms cannot be used, RPLSH is a good choice.  For example, RPLSH is very suitable for GPU, our study suggests the possibility of a GPU based ANNS algorithm which will outperform faiss (the current fastest GPU-based ANNS algorithm).
\end{itemize}

\section{Search with A Hash Index}

A hashing algorithm transforms the real vectors to binary codes. The binary codes can then be used as a hash index for online search. The general procedure of searching with a hash index is as follows (Suppose the user submit a query $q$ and ask for $K$ neighbors of the query):
\begin{enumerate}
	\item The search system encodes the query to a binary code $b$ use the hash functions.
	\item The search system finds $L$ points which are closest to $b$ in terms of the Hamming distance.
	\item The search system performs a scan within the $L$ points, returns the top $K$ points which are closet to $q$.
\end{enumerate}
This procedure is summarised in Algorithm \ref{alg:hashingsearch}, and there is one parameter $L$ which can be used to control the accuracy-efficiency trade-off. 
Based on this search procedure, we can see that the time of searching with a hash index contains three parts \cite{Jin2014Fast}:

\begin{algorithm}[t]
	\caption{Search with a hash index}
	\label{alg:hashingsearch}
	\begin{algorithmic}[1]
		\Require base set D, query vector $q$, the hash functions, hash index (binary codes for all the points in the base set), the pool size $L$ and  the number $K$ of required nearest neighbors,
		\Ensure $K$ points from the base set which are close to $q$.  
		\State Encode the query $q$ to a binary code $b$;
		\State Locate $L$ points which have shortest hamming distance to $b$;
		\State return the closet $K$ points to $q$ among these $L$ points.
	\end{algorithmic}
\end{algorithm}

\begin{enumerate}
	\item {\bf Coding} time $T_c$: the line 1 in Algorithm \ref{alg:hashingsearch}.
	\item {\bf Locating} time $T_l$: the lines 2 in Algorithm \ref{alg:hashingsearch}.
	\item {\bf Scanning} time $T_s$: the line 3 in Algorithm \ref{alg:hashingsearch}.
\end{enumerate}
The total search time $T = T_c+T_l+T_s$. For many popular hashing algorithms, $T_c$ can be neglected compared to $T_l$ and $T_s$, please see Table \ref{codingtime} for details. $T_s$ is not related to the hashing algorithm and only depends on the parameter $L$. Thus, correctly measuring $T_l$ becomes the key to evaluate different hashing algorithms. There are two popular procedures (the \textbf{hamming ranking} approach and the \textbf{hash bucket search} approach) to complete the task in line 2. And these two approaches need significantly different $T_l$ for different hashing algorithms on different data set.

In the remaining part of this Section, we will analyze the 
pros and cons of two search procedures using the hash index. We will also experimentally verify our analysis. The experiments are performed on two popular ANNS datasets, SIFT1M and GIST1M. The basic statistics of these two datasets are shown in Table \ref{data_detail_table}.

\begin{table}[t]\small
	\caption{Data sets}
	\label{data_detail_table}
	\centering
	\begin{tabular}{|c|c|c|c|}
		\hline
		data set &  dimension & base number & query number \\
		\hline
		SIFT1M &  128 & 1,000,000 & 10,000 \\
		GIST1M &  960 & 1,000,000 & 1,000 \\
		\hline
	\end{tabular}
\end{table}

We use the random-projection-based LSH \cite{Charikar2002LSH} for its simplicity. In the remaining part of this paper, the LSH algorithm we mentioned is this random-projection-based LSH.

Given a dataset with dimensionality $m$, one can generate an $m\times l$ random matrix from Gaussian distribution $\mathcal{N}(0,1)$ where $l$ is the required code length. Then the data will be projected onto $l$ dimensional space using this random matrix. The $l$ dimensional real representation of each point can then be converted to $l$-bits binary code by a simple binarize operation\footnote{$i$-th bit is 1 if $i$-th feature is greater or equal to 0; $i$-th bit is 0 if $i$-th feature is smaller than 0;}. The formal algorithm is illustrated in Algorithm (\ref{alg:RPLSH}).

Given a query point, the search algorithm is expected to return $K$ points. Then we need to examine how many points in this returned set are among the real $K$ nearest neighbors of the query. Suppose the returned set of $K$ points given a query is $R$ and the real $K$ nearest neighbors set of the query is $R'$, the $precision$ and $recall$ can be defined \cite{Makhoul2000Performance} as
\begin{equation}
precision(R) = \frac{|R' \cap R|}{|R|},\ \ \ recall(R) = \frac{|R' \cap R|}{|R'|}.
\end{equation}
Since the sizes of $R'$ and $R$ are the same, the recall and the precision of $R$ are actually the same. We fixed $K=100$ throughout our experiments.

\begin{algorithm}[t]
	\caption{random projection based LSH}
	\label{alg:RPLSH}
	\begin{algorithmic}[1]
		\Require base set D with $n$ $m$-dimensional points, code length $l$
		\Ensure $m\times l$ projection matrix $A$ and $n$ $l$-bits binary codes..  
		\State Generate an $m\times l$ random matrix $A$ from Gaussian distribution $\mathcal{N}(0,1)$;
		\State Project the data onto $l$ dimensional space using this random matrix. 
		\State Convert the $n\times l$ dimensional data matrix to n $l$-bits binary codes by the binarize operation.
	\end{algorithmic}
\end{algorithm}

\subsection{The Case That The Locating Time Being Ignored}

Suppose we have an ideal approach which can locate $L$ binary vectors closest (measured by the hamming distance) to the query at no cost, \ie, the locating time $T_l$ can be ignored, the {\bf \# of located samples}, $L$, becomes a good indicator of the search time.  Thus, the {\bf accuracy - \# of located samples} curve can used to compare the performance of different hashing algorithms. 

\begin{figure}[t]
	\centering
	\subfigure[SIFT1M]{\includegraphics[width=\QuaterFigureWidthSmall]{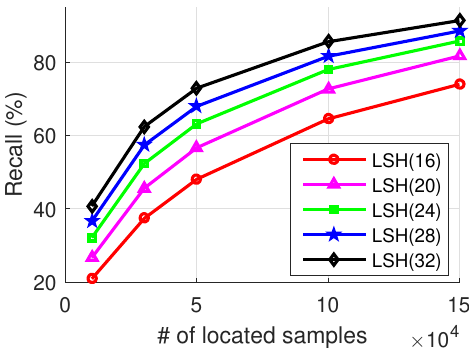}}
	\subfigure[GIST1M]{\includegraphics[width=\QuaterFigureWidthSmall]{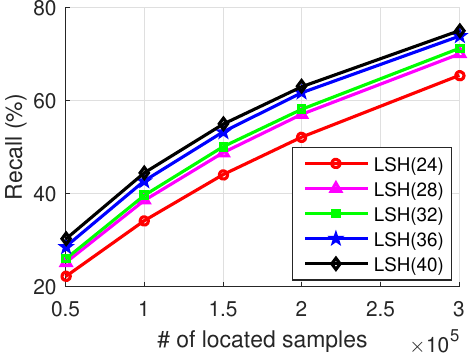}}
    \caption{The recall - \# of located samples curves of LSH on two datasets. We can see longer codes means better performance. }
	\label{PvsSamplevsLSH}
\end{figure}

Figure \ref{PvsSamplevsLSH} plots the {\bf recall - \# of located samples} curves of LSH on two datasets. The code length grows from 16 to 32 on SIFT1M and 24 to 40 on GIST1M. 
We can easily get the conclusion that longer code means better performance, which is a common conclusion in almost all the previous hashing papers \cite{WeissTF08NIPS,KulisG09ICCV,KulisD09NIPS,WangKC10ICML,ZhangWCL10SIGIR,HeLC10KDD,XuWLZLY11ICCV,LiuWKC11ICML,JegouDS11PAMI,LiuWJJC12CVPR,WeissFT12,LinJCYL13CVPR,HeWS13CVPR,GeHK013CVPR,JinHLZLCL13ICCV,XuBLCHC13IJCAI,JinLLC14TCYB,KalantidisA14CVPR,LiuMKC14NIPS}.

This result is very natural and reasonable. Since each bit of the hash code is a partition of the feature space and same code (0 or 1) on this bit means two points are on the same side of this partition. If two points share $t$ bits same code, which means these two points are on the same sides of $t$ partitions. Neighbors in the hamming space with longer code are more likely to be the real neighbors in the original feature space. However, this is only valid when the locating time is ignored.

\subsection{The Hamming Ranking Approach}

The hamming ranking approach is very simple and wildly used in many hashing papers \cite{WeissTF08NIPS,KulisG09ICCV,KulisD09NIPS,WangKC10ICML,ZhangWCL10SIGIR,HeLC10KDD,XuWLZLY11ICCV,LiuWKC11ICML,JegouDS11PAMI,LiuWJJC12CVPR,WeissFT12,LinJCYL13CVPR,HeWS13CVPR,GeHK013CVPR,JinHLZLCL13ICCV,XuBLCHC13IJCAI,JinLLC14TCYB,KalantidisA14CVPR,LiuMKC14NIPS}. The algorithm procedure of hamming ranking can be found in Algorithm \ref{alg:hammingRanking}.

\begin{algorithm}[t]
	\caption{Hamming ranking approach for locate $L$ closest binary codes}
	\label{alg:hammingRanking}
	\begin{algorithmic}[1]
		\State Compute the hamming distances between $b$ and all the binary codes in the data set;
		\State Sort the points based on the hamming distances in ascending order and keep the first $L$ points;
	\end{algorithmic}
\end{algorithm}

It is easy to find that the computational complexity of this approach is $O(N)$ where $N$ is the data points. Thus, the reason of using hash index (hamming ranking approach) to speed up search is that the computation of hamming distance of two binary codes can be extremely fast \cite{Hacker}.

\begin{figure}[t]
	\centering
	\subfigure[]{\includegraphics[width=\QuaterFigureWidthSmall]{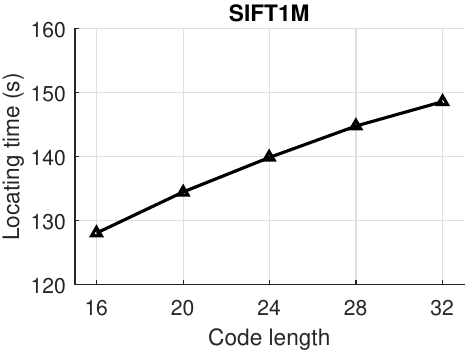}
				\includegraphics[width=\QuaterFigureWidthSmall]{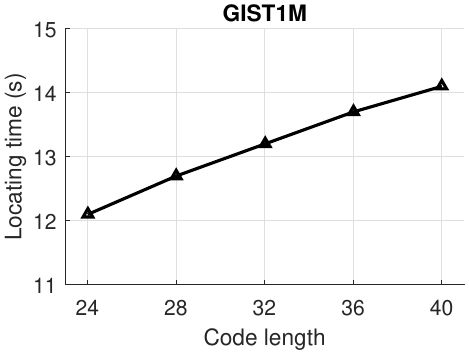}}
	\subfigure[]{\includegraphics[width=\QuaterFigureWidthSmall]{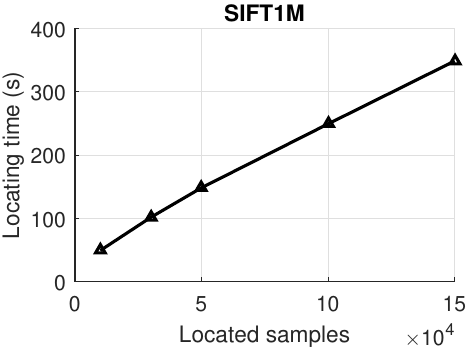}
		\includegraphics[width=\QuaterFigureWidthSmall]{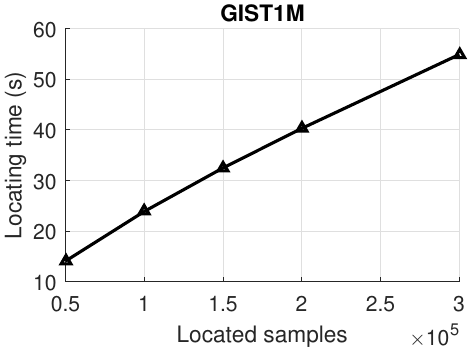}}
	\caption{The locating time complexity of the hamming ranking approach. (a) The locating time VS. the code length ($L=50,000$) (b) The locating time VS. the parameter $L$. The locating time grows linearly with respect to both the code length and the parameter $L$. }
	\label{LocatingHamming}
\end{figure}

The locating time of hamming ranking approach (the time of Algorithm \ref{alg:hammingRanking}) is only determined by the code length and the parameter $L$. In other words, if two different hashing algorithms use the same code length and the same $L$, they will have the same locating time. Thus, we can compare different hashing algorithms with the same code length simply using the {\bf accuracy - \# of located samples} curves (ignore the locating time). This strategy is used in most of the exiting hashing papers \cite{WeissTF08NIPS,KulisG09ICCV,KulisD09NIPS,WangKC10ICML,ZhangWCL10SIGIR,HeLC10KDD,XuWLZLY11ICCV,LiuWKC11ICML,JegouDS11PAMI,LiuWJJC12CVPR,WeissFT12,LinJCYL13CVPR,HeWS13CVPR,GeHK013CVPR,JinHLZLCL13ICCV,XuBLCHC13IJCAI,JinLLC14TCYB,KalantidisA14CVPR,LiuMKC14NIPS}. Moreover, The locating time grows linearly with respect to the code length and the parameter $L$ (as shown in Figure \ref{LocatingHamming}).

\begin{figure}[t]
	\centering
	\subfigure[SIFT1M]{\includegraphics[width=\QuaterFigureWidthSmall]{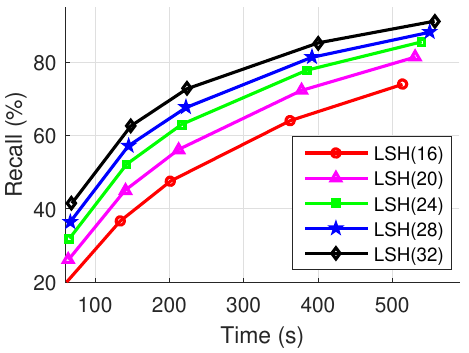}}
	\subfigure[GIST1M]{\includegraphics[width=\QuaterFigureWidthSmall]{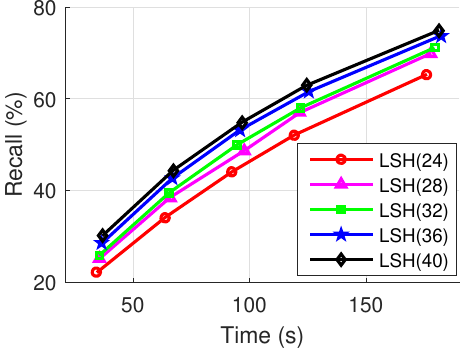}}
	\caption{The recall - search time curves of LSH using hamming ranking approach on two datasets.}
	\label{PvsTimeHamming}
\end{figure}

\begin{figure}[t]
	\centering
	\subfigure[]{\includegraphics[width=\QuaterFigureWidthSmall]{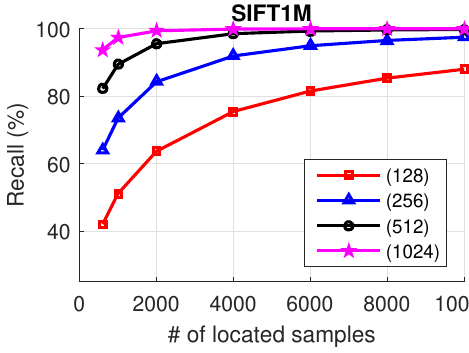}}
	\subfigure[]{\includegraphics[width=\QuaterFigureWidthSmall]{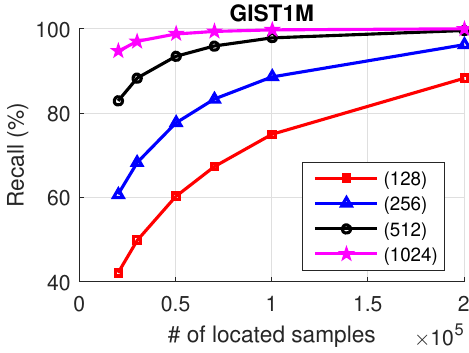}}
	\subfigure[]{\includegraphics[width=\QuaterFigureWidthSmall]{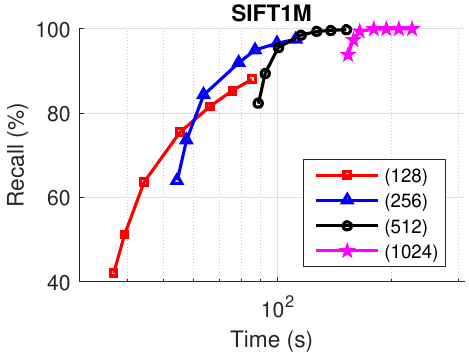}}
	\subfigure[]{\includegraphics[width=\QuaterFigureWidthSmall]{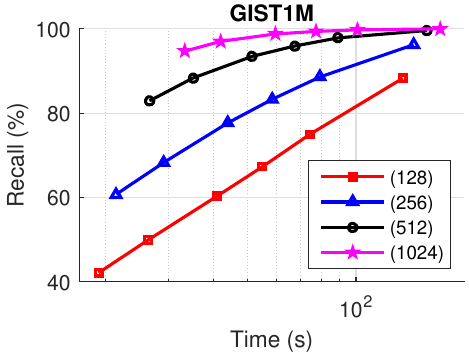}}
	\caption{The ANN search results of LSH with significant different code length on two datasets. The y-axis denotes the average recall while the x-axis denotes (a,b) the \# of located samples (c,d) the search time using the hamming ranking approach.}
	\label{PvsTimeHamming2}
\end{figure}

 Figure \ref{PvsTimeHamming} shows the {\bf recall - search time} curves of LSH using hamming ranking approach on two datasets. This result is very similar to the result in Figure \ref{PvsSamplevsLSH}. It is because the locating time grows slowly as the code length increases and the code length differences between different curves in Figure \ref{PvsTimeHamming} are also small. As the code length difference becomes larger, we can see the impact of the locating time, as shown in Figure \ref{PvsTimeHamming2}.

One advantage of the hamming ranking approach is that the memory usage of the index is extremely small. From Algorithm \ref{alg:hammingRanking}, we can find that there is no additional data structure needed and we only need to store the binary codes for the data points in the database. Suppose the database contains $n$ data points and we use $l$ bits code, we need $n\frac{l}{8}$ byte. This memory usage may be the smallest one among most of the popular ANNS methods even when $l=1024$.

\subsection{The Hash Bucket Search Approach}

\begin{table*}[t]\scriptsize
	\caption{Hamming radius VS. \# of queries successfully located 50000 points on SIFT1M (total \# of queries 10,000)}
	\label{hammingRadiusSIFT}
	\centering
	\begin{tabular}{|c|c|c|c|c|c|c|c|c|c|c|c|}
		\hline
		hamming radius &  0 & 1 & 2 & 3 & 4 &  5 & 6 & 7 & 8 & 9 &10\\
		\hline
		\hline
		\# of buckets (16 bits) &  1 & 16 & 120 & 560 & 1,820 &  4,368 & 8,008 & 11,440 & 12,870 & 11,440 & 8,008\\
		\hline
		\# of queries (16 bits) (1.752s) &  0 & 3637 & 4932 & 1230 & 169 &  30 & 1 & 1 & 0 & 0 & 0\\
		\hline
		\hline
		\# of buckets (32 bits) &  1 & 32 & 496 & 4960 & 35,960 &  201,376 & 906,192 & 3,365,856 & 10,518,300 & 28,048,800 & 64,512,240\\
		\hline
		\# of queries (32 bits) (237.8s) &  0 & 0 & 0 & 4 & 1710 &  4282 & 2691 & 1021 & 241 & 44 & 7\\
		\hline
	\end{tabular}
\end{table*}

\begin{table*}[t]\scriptsize
	\caption{Hamming radius VS. \# of queries successfully located 50000 points on GIST1M (total \# of queries 1,000)}
	\label{hammingRadiusGIST}
	\centering
	\begin{tabular}{|c|c|c|c|c|c|c|c|c|c|c|c|}
		\hline
		hamming radius &  0 & 1 & 2 & 3 & 4 &  5 & 6 & 7 & 8 & 9 &10\\
		\hline
		\hline
		\# of buckets (24 bits) &  1 & 24 & 276 & 2,024 & 10,626 &  42,504 & 134,596 & 346,104 & 735,471 & 1,307,504 & 1,961,256\\
		\hline
		\# of queries (24 bits) (0.234s) &  70 & 467 & 277 & 115 & 44 &  18 & 7 & 2 & 0 & 0 & 0\\
		\hline
		\hline
		\# of buckets (40 bits) &  1 & 40 & 780 & 9,880 & 91,390 &  658,008 & 3,838,380 & 18,643,560 & 76,904,685 & 273,438,880 & 847,660,528\\
		\hline
		\# of queries (40 bits) (240.0s) &  0 & 0 & 27 & 252 & 302 &  205 & 107 & 57 & 27 & 14 & 9\\
		\hline
	\end{tabular}
\end{table*}

\begin{algorithm}[t]
	\caption{Hash bucket search approach for locate $L$ closest binary codes}
	\label{alg:hashBucket}
	\begin{algorithmic}[1]
		\State Let hamming radius $r = 0$;
		\State Let candidate pool set $C = \emptyset$;
		\While{$|C| < L$}
		\State Get the buckets list $B$ with hamming distance $r$ to $b$;
		\ForAll{bucket $bb$ in $B$}
		\State Put all the points in $bb$ into $C$;
		\If {$|C| >= L$}
		\State Keep the first $L$ points in $C$;
		\State {\bf return};
		\EndIf
		\EndFor
		\State $r = r+1$;
		\EndWhile
	\end{algorithmic}
\end{algorithm}

\begin{figure}[t]
	\centering
	\subfigure[SIFT1M]{\includegraphics[width=\QuaterFigureWidthSmall]{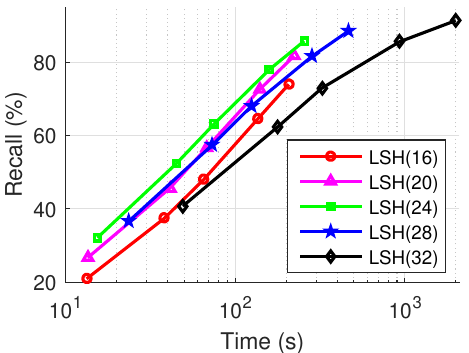}}
	\subfigure[GIST1M]{\includegraphics[width=\QuaterFigureWidthSmall]{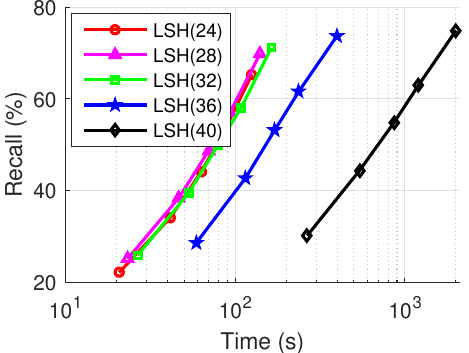}}
	\caption{The recall - search time curves of LSH using hash bucket search  approach on two datasets.}
	\label{PvsTimeHash}
\end{figure}

\begin{figure}[t]
	\centering
	\subfigure[]{\includegraphics[width=\QuaterFigureWidthSmall]{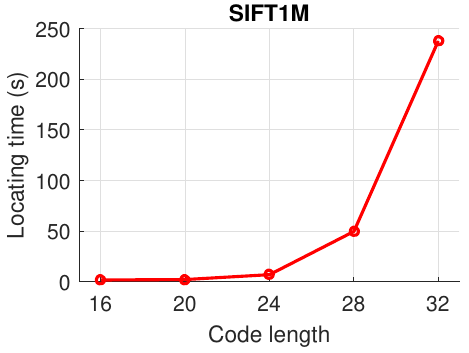}
		\includegraphics[width=\QuaterFigureWidthSmall]{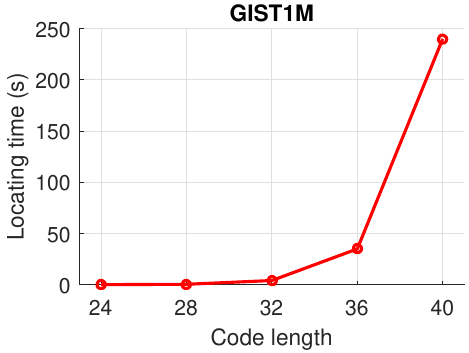}}
	\subfigure[]{\includegraphics[width=\QuaterFigureWidthSmall]{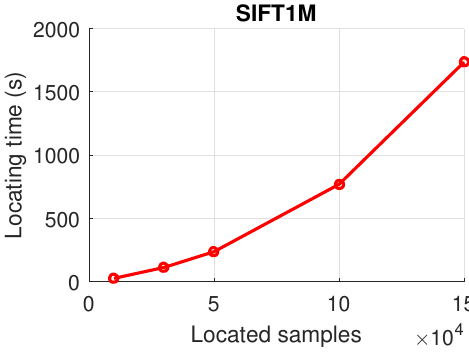}
		\includegraphics[width=\QuaterFigureWidthSmall]{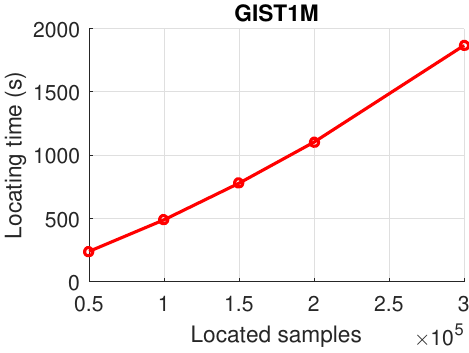}}
	\caption{The locating time complexity of the hash bucket search approach. (a) The locating time VS. the code length ($L=50,000$) (b) The locating time VS. the parameter $L$. The locating time grows exponentially with respect to the code length and grows super-linearly with the parameter $L$. }
	\label{LocatingHashing}
\end{figure}

The hash bucket search is another popular method to locate those nearest binary codes in the hamming space. The corresponding algorithm procedure can be found in Algorithm \ref{alg:hashBucket}.

Figure \ref{PvsTimeHash} shows the {\bf recall - time} curves using the hash bucket search approach of LSH on two datasets. These curves are significant different with the curves in Figure \ref{PvsSamplevsLSH}
and Figure \ref{PvsTimeHamming}. There is an optimal code length for hash bucket search approach (24 in SIFT1M and 28 in GIST1M). As the code length further increases, the search performance decreases significantly.

Given a binary code $b$, locating the hashing bucket corresponding to $b$ costs $O(1)$ time (by using std::vector). If there are enough points (larger than $L$) in this bucket, the total time cost of the hash bucket search is $O(1)$, which seems much more efficient than the hamming ranking approach. However, this is only true when the binary code is short. As the code length increases, the number of hash bucket increases exponentially. And there are always not enough points (even no point) in the bucket corresponding to the query. To locate $L$ points, We need to increase the hamming radius $r$. Given an $l$-bits  code $b$, considering those hashing buckets whose Hamming distance to $b$ is smaller or equal to $r$. It is easy to show the number of these buckets is $\sum_{i=0}^r\binom{l}{i}$, which increases almost exponentially with respect to $r$ and $l$. As the $L$ increases, to locate $L$ points successfully, $r$ has to be increased. Overall, the locating time of the hash bucket search approach grows {\bf exponentially} with respect to the code length $l$ and super-linearly with respect to the parameter $L$ (as shown in Figure \ref{LocatingHashing}). 

Table \ref{hammingRadiusSIFT} and \ref{hammingRadiusGIST} record the number of queries (the total number is 10,000 on SIFT1M and 1,000 on GIST1M) which successfully located 50,000 samples in different hamming radius with different code length with the  hash bucket search approach. As the code length increases, the number of hash buckets corresponding to the same hamming radius increases almost exponentially. Moreover, as the code length increases, the base points are more diversified. Thus we need to increase the hamming radius to locate enough number of samples. These two reasons cause the locating time grows exponentially as the code length increases for the hash bucket search approach. As a result, the hash bucket search approach has no practical meaning if the code length is longer than 64.

However, Section 2.1 shows long codes can generate better search candidates. A natural question will be can we design a method for the hash bucket search approach to take the advantages of long codes?

The answer is yes. A straightforward way is using multiple hash tables to represent long codes. Suppose we have 128-bits codes and we use a single table with 128-bits. To locate $L$ points, if the points in all the buckets within the Hamming distance $r$ are not enough, we have to increase the hamming radius $r$.  The locating time will then grow significantly and the extremely long locating time will make the hash bucket search approach meaningless. If we use four tables, each table is 32-bits. Instead of growing $r$, we can scan the buckets within the hamming radius $r$ in all the tables, which gives us a more substantial chance to locate enough data points.

\begin{figure}[t]
	\centering
	\subfigure[SIFT1M]{\includegraphics[width=\QuaterFigureWidthSmall]{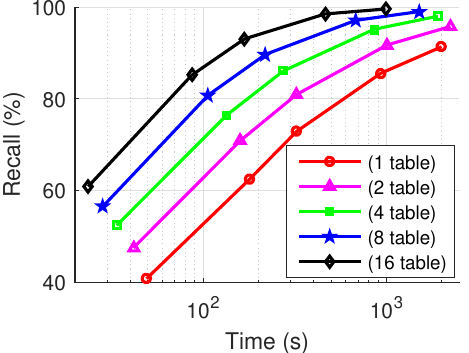}}
	\subfigure[GIST1M]{\includegraphics[width=\QuaterFigureWidthSmall]{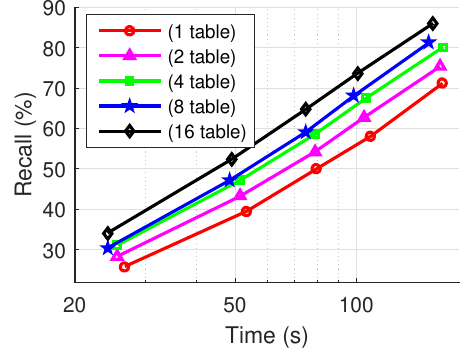}}
	\caption{The {\bf recall-time} curves of the hash bucket search approach with different number of tables on SIFT1M and GIST1M.}
	\label{nTableLSH}
\end{figure}

Figure \ref{nTableLSH} shows the {\bf recall-time} curves of the hash bucket search approach on LSH code with the different number of tables (each table is 32-bits). The results are very consistent: the hash bucket search approach gains the advantage by using multiple tables on both datasets. This multiple tables trick is more effective on SIFT1M than on GIST1M. 

With 1,024 bits code, we can use $\lceil\frac{1024}{w} \rceil$ tables, each table is $w \leq 1024$ bits. If we use the {\bf recall - time} curve,  the performance of the hash bucket search approach will firstly increase then dramatically decrease as the $w$ increases. There will be an optimal $w$ given a hashing algorithm and the dataset.

\begin{figure}
	\centering
	\subfigure[SIFT1M]{\includegraphics[width=\DoubleFigureWidth]{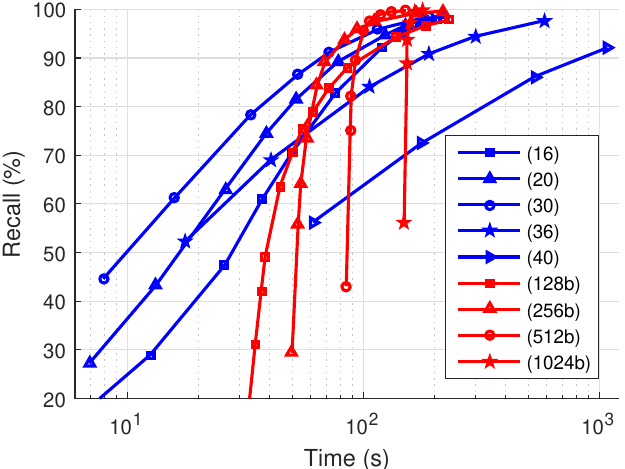}}
	\subfigure[GIST1M]{\includegraphics[width=\DoubleFigureWidth]{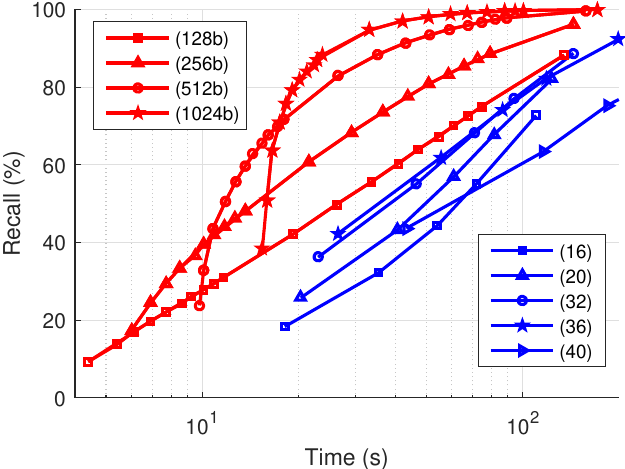}}
	\caption{The {\bf recall - time} curves of LSH with the hamming ranking and the hash bucket search approaches on SIFT1M and GIST1M. The read curves are the hamming ranking approach and the blue curves are the hash bucket search approach. The hamming ranking approach uses 128, 256, 512 and 1,024 bits and the hash bucket search approach use $\lceil\frac{1024}{w} \rceil$ tables to represent 1024 bits code, each table is $w$ bits, where table width $w$ is 16, 20, 30(32), 36 and 40.  }
	\label{LSHHashingVsHamming}
\end{figure}

Figure \ref{LSHHashingVsHamming} shows the performance of LSH on two datasets with both hamming ranking and hash bucket search approaches. We use 128, 256, 512 and 1024 bits for the hamming ranking approach. For the hash bucket search approach, we use $\lceil\frac{1024}{w} \rceil$ tables to represent 1024 bits code,  each table is $w$ bits, where table width $w$ is 16, 20, 30(32), 36 and 40. 

Generally speaking, the hash bucket search approach has the advantage on SIFT1M while the hamming ranking approach is preferred on GIST1M. For the hash bucket search approach, the optimal $m$ is 34 on SIFT1M and 28 on GIST1M. For the hamming ranking approach, longer code is preferred when we aim at a high recall while short code has the advantage when we aim at a low recall. It is important to note that these optimal parameters  (the code length for the hamming ranking and table number for the hash bucket search) will change if we use another hashing algorithm (see Figure \ref{SIFT_Ghamming} and \ref{GIST_Ghamming}). A fair comparison between various hashing algorithms seems impossible.

\begin{figure*}
	\centering
	\subfigure[SIFT1M]{\includegraphics[width=\DoubleFigureWidthTiny]{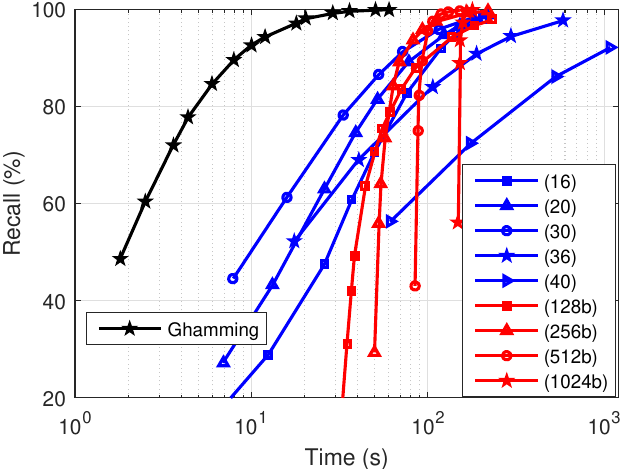}}
	\subfigure[GIST1M]{\includegraphics[width=\DoubleFigureWidthTiny]{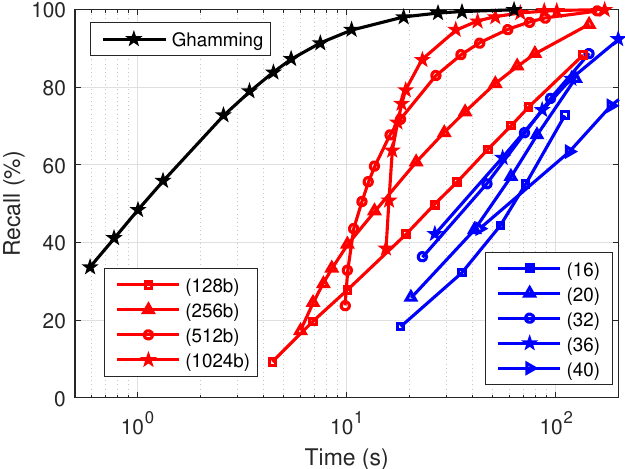}}
	\caption{The {\bf recall-time} curves of  LSH with the grouped hamming ranking, the hamming ranking and the hash bucket search approaches on SIFT1M and GIST1M. The red and blue curves are the same as the curves in Figure \ref{LSHHashingVsHamming}.  The black curve is the grouped hamming ranking approach proposed in this paper. The grouped hamming ranking approach is significant better than other two approaches on both datasets.}
	\label{HashingVsHamming}
\end{figure*}

\section{Grouped Hamming Ranking Approach}

\begin{figure*}[t]
	\centering
	\includegraphics[width=\QuaterFigureWidth]{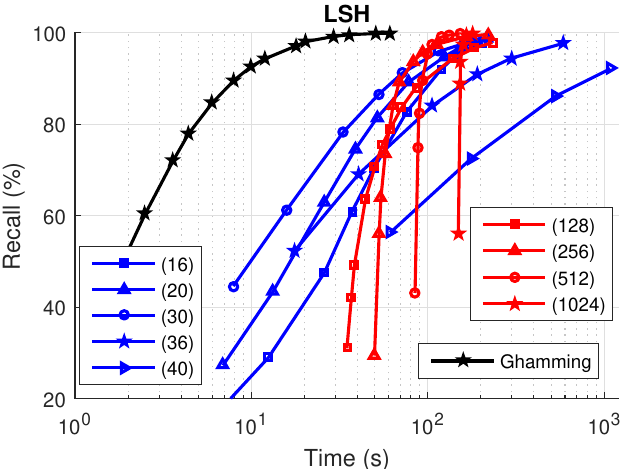}
	\includegraphics[width=\QuaterFigureWidth]{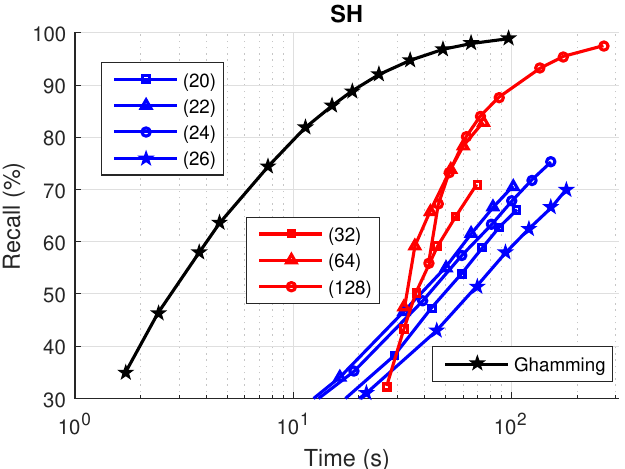}
	\includegraphics[width=\QuaterFigureWidth]{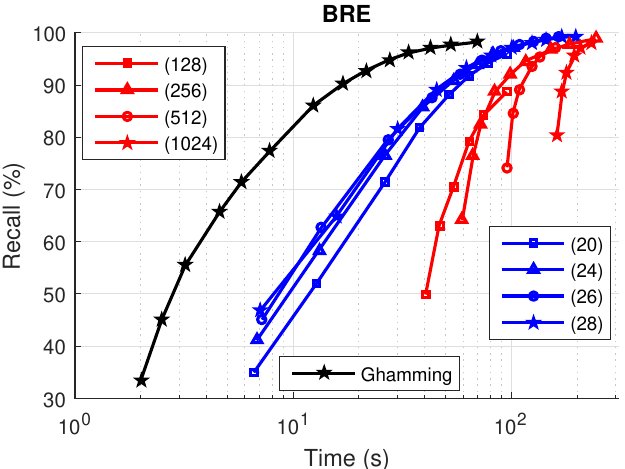}
	\includegraphics[width=\QuaterFigureWidth]{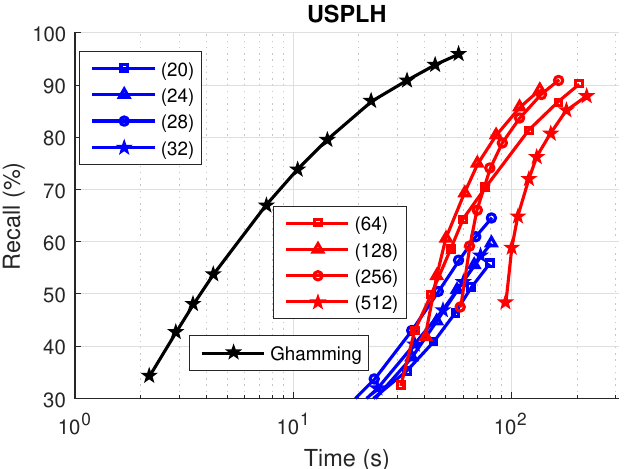}
	\includegraphics[width=\QuaterFigureWidth]{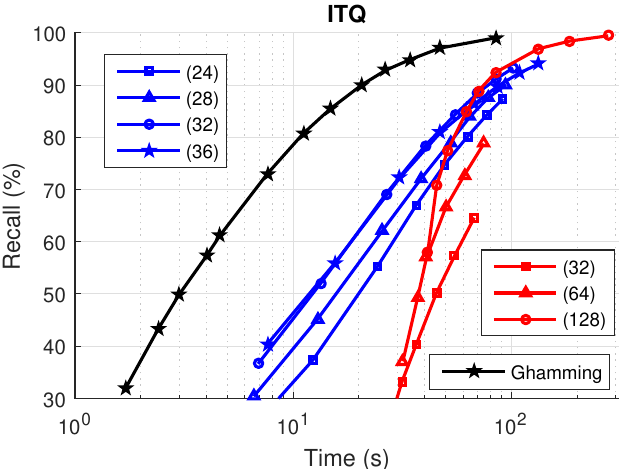}
	\includegraphics[width=\QuaterFigureWidth]{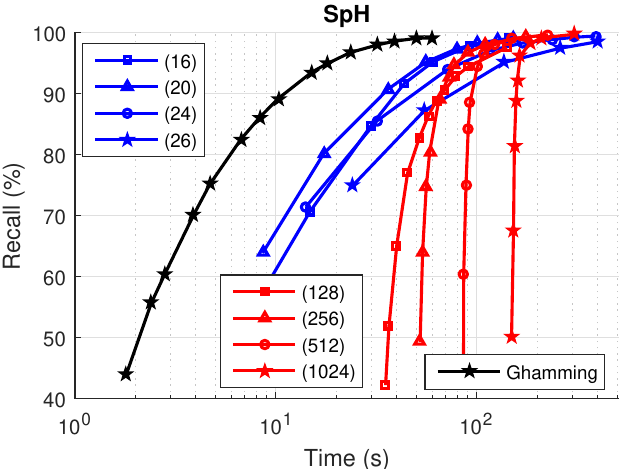}
	\includegraphics[width=\QuaterFigureWidth]{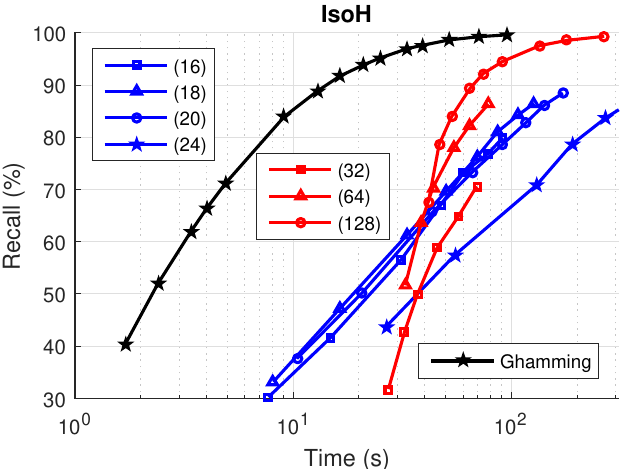}
	\includegraphics[width=\QuaterFigureWidth]{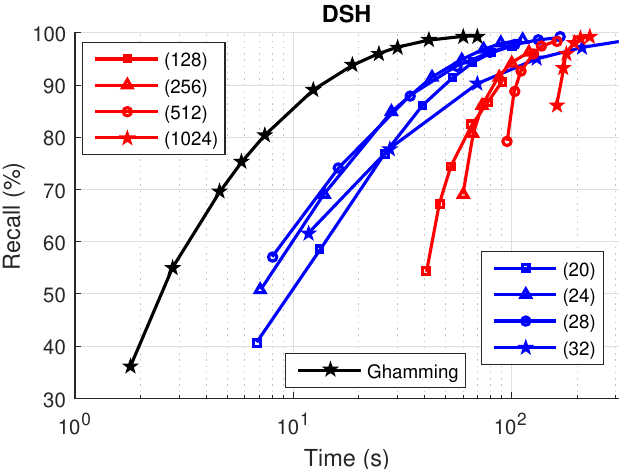}
	\includegraphics[width=\QuaterFigureWidth]{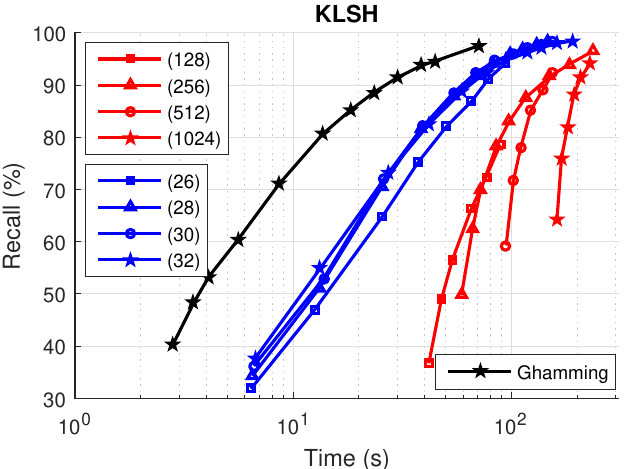}
	\includegraphics[width=\QuaterFigureWidth]{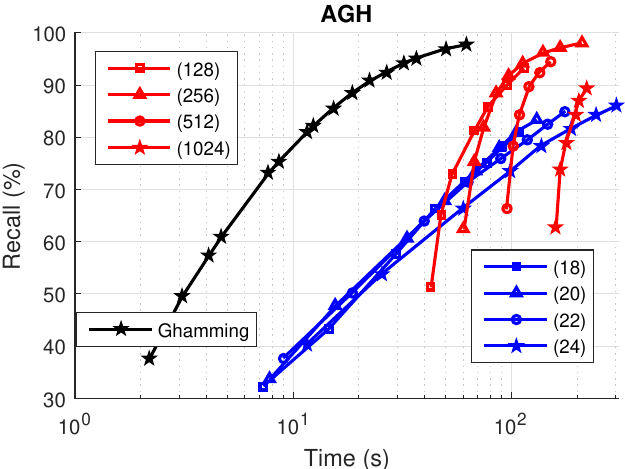}
	\includegraphics[width=\QuaterFigureWidth]{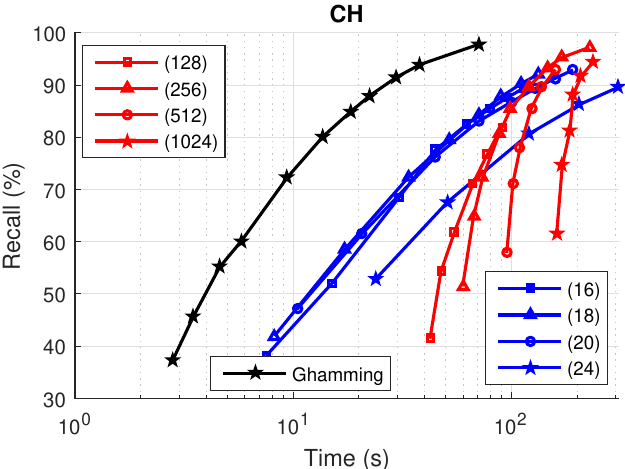}
	\caption{The comparison of three search-the-hash-index approaches with 11 hashing algorithms on SIFT1M. The red curves are hamming ranking approaches with the numbers in the legend indicate the code length. The blue curves are hash bucket search approaches with the numbers in the legend indicates the table width. If the $w$ is the table width, the corresponding table number is $\lceil\frac{1024}{w} \rceil$. The black curve is the grouped hamming ranking approach proposed in this paper. No matter which hashing algorithm is used, the grouped hamming ranking approach is significant better than other two approaches. }
	\label{SIFT_Ghamming}
\end{figure*}

\begin{figure*}[t]
	\centering
	\includegraphics[width=\QuaterFigureWidth]{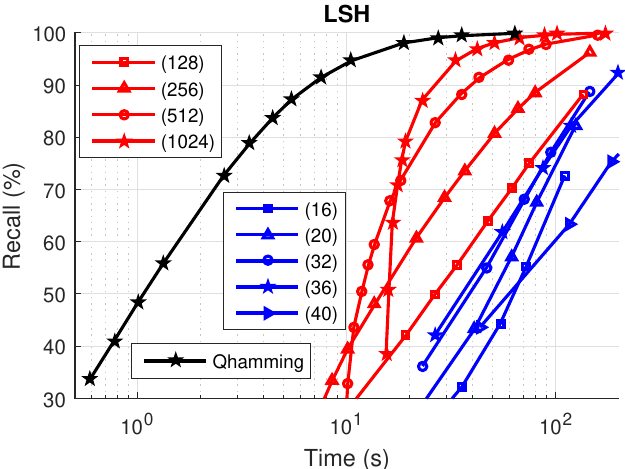}
	\includegraphics[width=\QuaterFigureWidth]{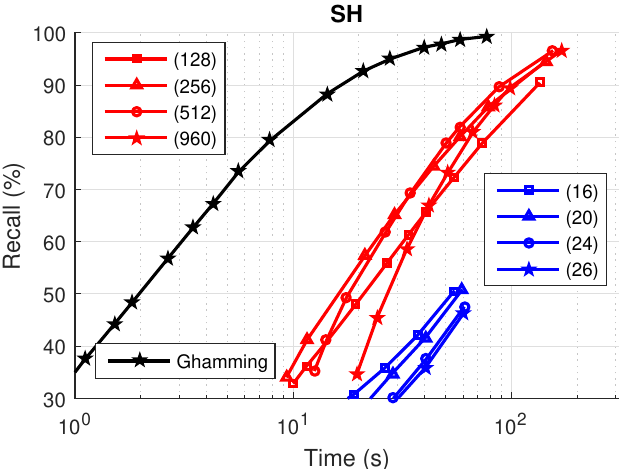}
	\includegraphics[width=\QuaterFigureWidth]{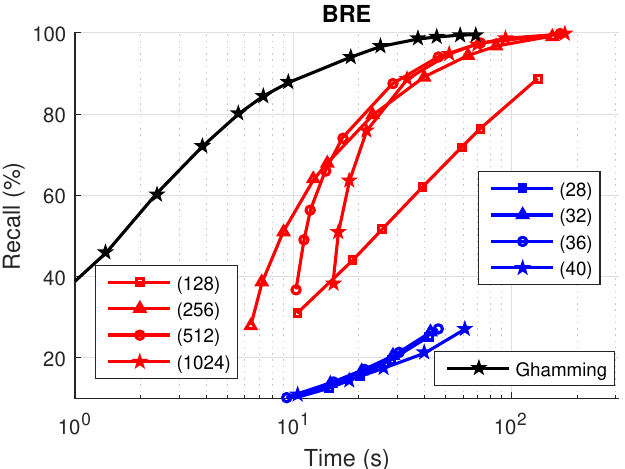}
	\includegraphics[width=\QuaterFigureWidth]{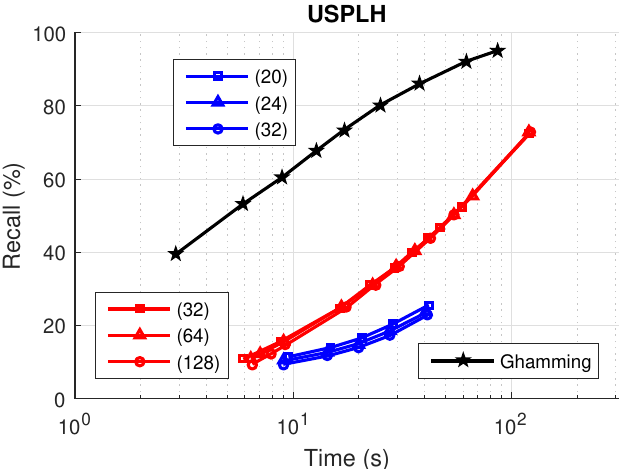}
	\includegraphics[width=\QuaterFigureWidth]{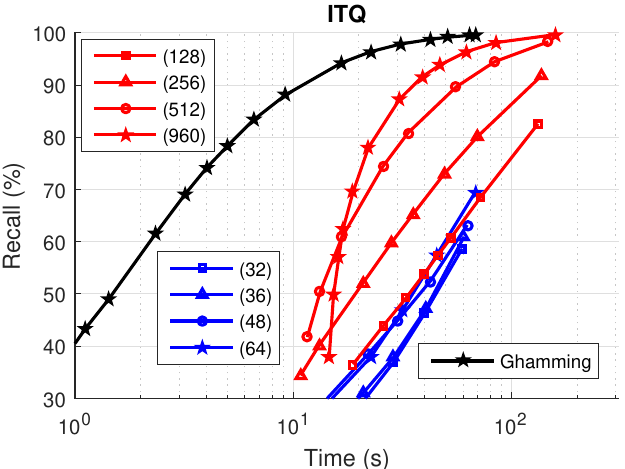}
	\includegraphics[width=\QuaterFigureWidth]{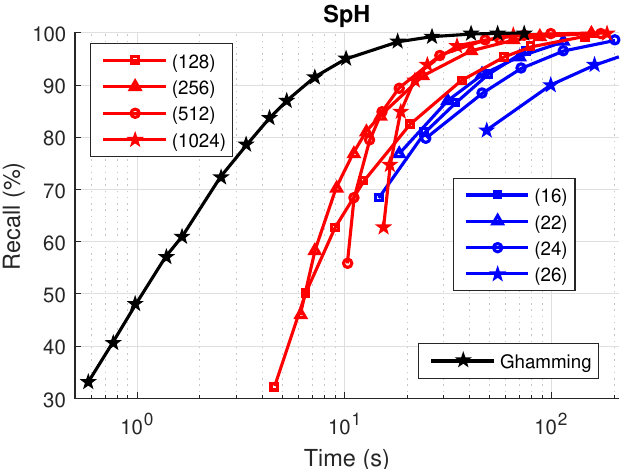}
	\includegraphics[width=\QuaterFigureWidth]{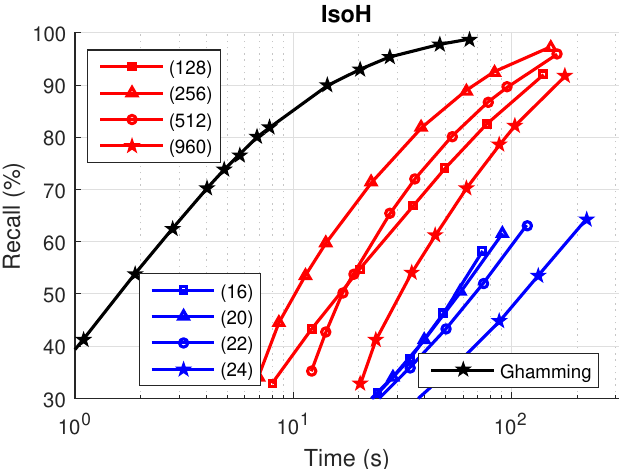}
	\includegraphics[width=\QuaterFigureWidth]{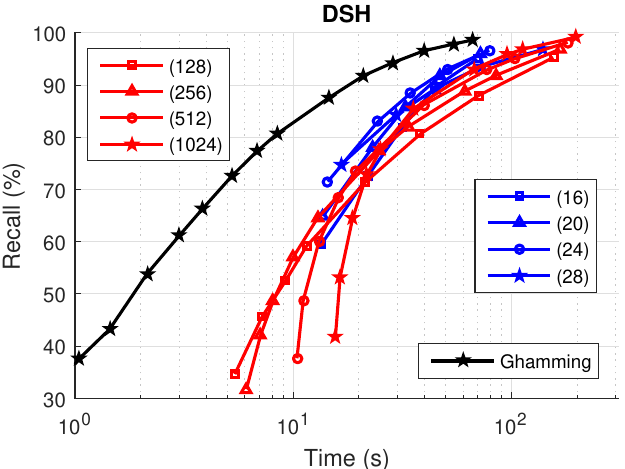}
	\includegraphics[width=\QuaterFigureWidth]{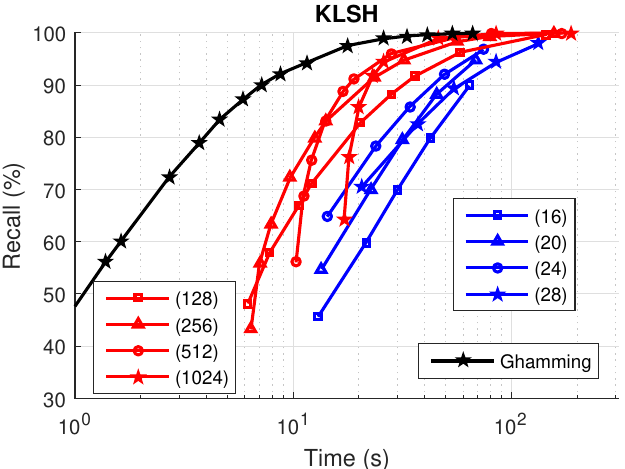}
	\includegraphics[width=\QuaterFigureWidth]{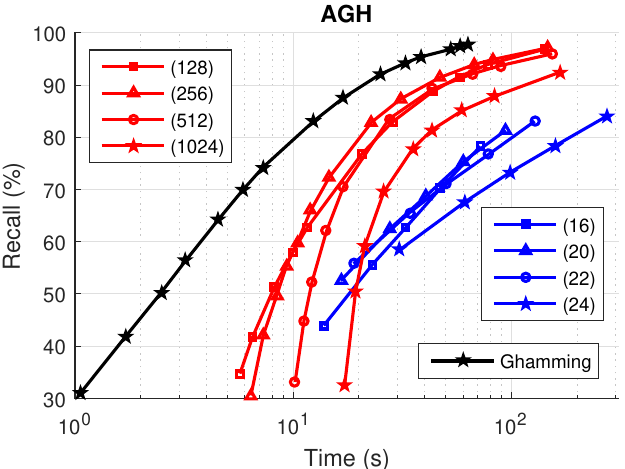}
	\includegraphics[width=\QuaterFigureWidth]{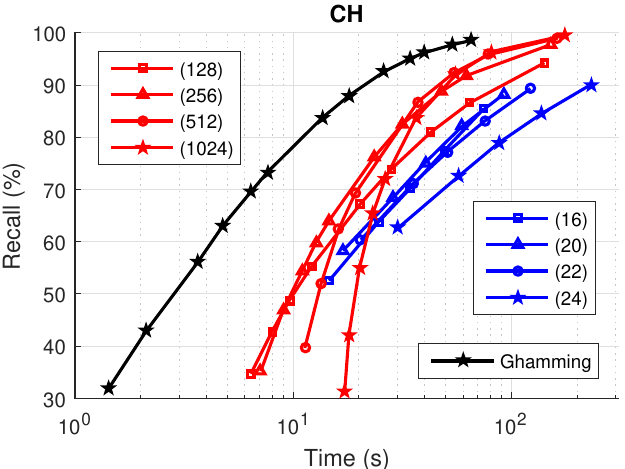}
	\caption{The comparison of three search-the-hash-index approaches with 11 hashing algorithms on GIST1M. The red curves are hamming ranking approaches with the numbers in the legend indicate the code length. The blue curves are hash bucket search approaches with the numbers in the legend indicates the table width. If the $w$ is the table width, the corresponding table number is $\lceil\frac{1024}{w} \rceil$. The black curve is the grouped hamming ranking approach proposed in this paper. No matter which hashing algorithm is used, the grouped hamming ranking approach is significant better than other two approaches. }
	\label{GIST_Ghamming}
\end{figure*}

The commonly used approaches (the hamming ranking and the hash bucket search) for search with a hash index have their own pros and cons. Different hashing algorithms prefer different approach on different dataset (see Figure \ref{SIFT_Ghamming} and \ref{GIST_Ghamming}). It is not a easy task to fairly compare different hashing algorithms simply based on these two approaches. In this section, we are aiming at developing a better way to search with a hash index.

\begin{algorithm}[t]
	\caption{Grouped hamming ranking}
	\label{alg:Qhamming}
	\begin{algorithmic}[1]
		\Require base set D, query vector $q$, the hash functions, hash index (binary codes for all the points in the base set), kmeans partition of the base set, the nearest clusters $C$,the pool size $L$ and the number $K$ of required nearest neighbors,
		\Ensure $K$ points from the base set which are close to $q$.  
		\State Encode the query $q$ to a binary code $b$;
		\State Find the $C$ nearest clusters to $q$ 
		\State Compute the hamming distances between $b$ and all the binary codes in $C$ clusters;
		\State Find the first $L$ points with shortest hamming distance to $b$;Grouped
		\State return the closet $K$ points to $q$ among these $L$ points.
	\end{algorithmic}
\end{algorithm}

The proposed approach is based on the hamming ranking approach. Recall that the majority part of the locating time using the hamming ranking approach spends on the computation of the Hamming distances of the query code and all the database codes. If we can find an efficient wa to filter out those points which are unlikely to be the answers, we can reduce this computation.   

Follow the the idea in \cite{JegouDS11PAMI}, we use kmeans to group the samples into $k$ clusters at the indexing stage, and we represent each cluster with its centroid. At the on-line query stage, instead of computing the Hamming distances of the query code and the entire database codes, we can only focus on the nearest $c$ clusters (measured by the distance of the query vector and the centroid). We named this approach as {\bf Grouped Hamming Ranking}.

To use grouped hamming ranking, besides the binary code, we need a kmeans partition of the database. The search procedure is summarized in Algorithm \ref{alg:Qhamming}. There will be two parameters $C$ and $L$ which can be used to control the accuracy-efficiency trade-off.

Figure \ref{HashingVsHamming} shows the performance of LSH on two datasets with the grouped hamming ranking approach (we generated 1,000 clusters with kmeans and 1,024 bits codes were used). The curve of grouped hamming ranking approach is added on the Figure \ref{LSHHashingVsHamming} and we can see the significant advantage of the grouped hamming ranking approach over the other two commonly used approaches.  Figure \ref{SIFT_Ghamming} and \ref{GIST_Ghamming} show the performance of three search-the-hash-index approaches with 11 hashing algorithms on two datasets. The grouped hamming ranking approach outperforms the other two approaches for all the hashing algorithms on both two datasets.

\section{A Comprehensive Comparison}

With the proposed grouped hamming ranking approach, we are ready to conduct a fair comparison between different hashing algorithms.
In this Section, we will conduct a comprehensive comparison between eleven popular hashing algorithms on SIFT1M and GIST1M.

\subsection{Compared Algorithms}

Eleven popular hashing algorithms compared in the experiments are listed as follows:
\begin{itemize}
	\item {\bf LSH} is a short name for random-projection-based Locality Sensitive Hashing \cite{Charikar2002LSH} as in Algorithm \ref{alg:RPLSH}. It is frequently used as a baseline method in various hashing papers.
	\item {\bf SH} is a short name for Spectral Hashing \cite{WeissTF08NIPS}. SH is based on quantizing
	the values of analytical eigenfunctions computed along PCA directions of the data.
	\item {\bf KLSH} is a short name for Kernelized Locality Sensitive Hashing \cite{KulisG09ICCV}. KLSH generalizes the LSH method to the kernel space.
	\item {\bf BRE} is a short name for  Binary Reconstructive Embeddings \cite{KulisD09NIPS}.
	\item {\bf USPLH} is a short name for Unsupervised Sequential Projection Learning Hashing \cite{WangKC10ICML}.
	\item {\bf ITQ} is a short name for ITerative Quantization \cite{GongL11CVPR}. ITQ finds a rotation
	of zero-centered data so as to minimize the quantization error of mapping this data to the vertices of a zero-centered
	binary hypercube.
	\item {\bf AGH} is a short name for Anchor Graph Hashing \cite{LiuWKC11ICML}. It aims at performing spectral analysis \cite{Belkin01Eigenmap} of the data which shares the same goal with Self-taught Hashing \cite{ZhangWCL10SIGIR}. The advantage of AGH over Self-taught Hashing is the computational efficiency. AGH uses an anchor graph \cite{Liu10AnchorGraph} to speed up the spectral analysis. 
	
\begin{table}[t]\scriptsize
	\caption{Indexing Time (s)}
	\label{indexingtime}
	\centering
	\begin{tabular}{|c|c|c||c|c|c|}
		\hline
		Method &  SIFT1M & GIST1M & Method &  SIFT1M & GIST1M \\
		\hline
		LSH &  3.35 & 12.33 & SH & 83.44 & 3864.3 \\
		BRE &  680.3 & 369.49 & IsoH & 1.68 & 91.99 \\
		USPLH & 1099.2 & 10508.3 & KLSH & 94.09 & 194.89 \\
		ITQ &  88.3 & 1372.4 & AGH & 296.84  & 389.91 \\
		SpH &  2583.5 & 4041.1 & CH & 263.78 & 365.34 \\
		DSH & 51.92 & 122.84 & & &  \\
		\hline
		\multicolumn{6}{l}{ITQ, SH and IsoH learn 128 bits code on SIFT1M and 960 bits code}\\
		\multicolumn{6}{l}{on GIST1M.}\\
	\end{tabular}
\end{table}

\begin{table}[t]\scriptsize
	\caption{Coding Time (s)}
	\label{codingtime}
	\centering
	\begin{tabular}{|c|c|c||c|c|c|}
		\hline
		Method &  SIFT1M & GIST1M & Method &  SIFT1M & GIST1M \\
		\hline
		LSH &  0.03 & 0.016 & SH & 1.21 & 6.82 \\
		BRE &  0.25 & 0.043 & IsoH & 0.007 & 0.034 \\
		USPLH &  0.05 & 0.02 & KLSH & 0.34 & 0.06 \\
		ITQ &  0.01 & 0.025 & AGH & 0.86  & 0.11 \\
		SpH &  0.11 & 0.027 & CH & 0.96 & 0.10 \\
		DSH &  0.04 & 0.02 & & &  \\
		\hline
		\multicolumn{6}{l}{ITQ, SH and IsoH use 128 bits code on SIFT1M and 960 bits code}\\
		\multicolumn{6}{l}{on GIST1M.}\\
	\end{tabular}
\end{table}

\begin{figure*}
	\centering
	\includegraphics[width=\SmallDoubleFigureWidth]{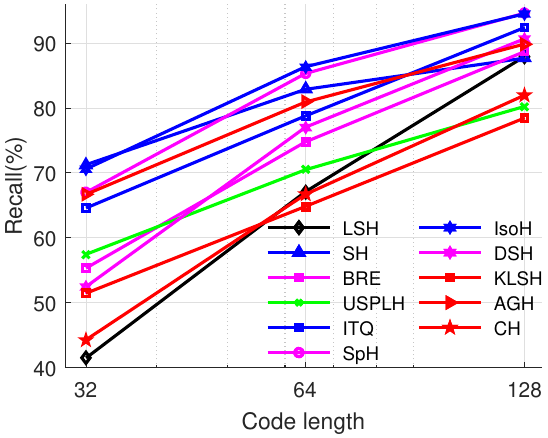}\includegraphics[width=\SmallDoubleFigureWidth]{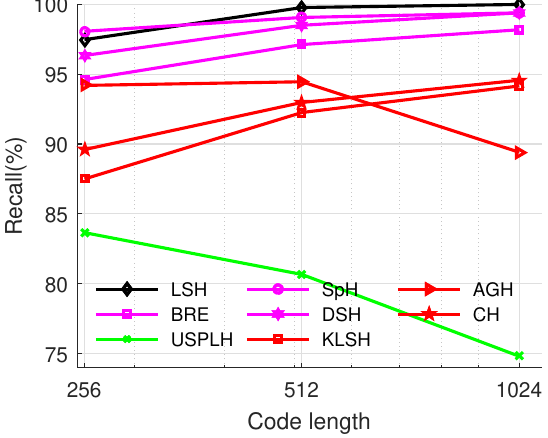}
	\caption{The {\bf recall - codelength} curves of various hashing algorithms when locate 10000 samples on SIFT1M}
	\label{PvsCodelengthSIFT}
\end{figure*}

\begin{figure*}
	\centering
	\includegraphics[width=\SmallDoubleFigureWidth]{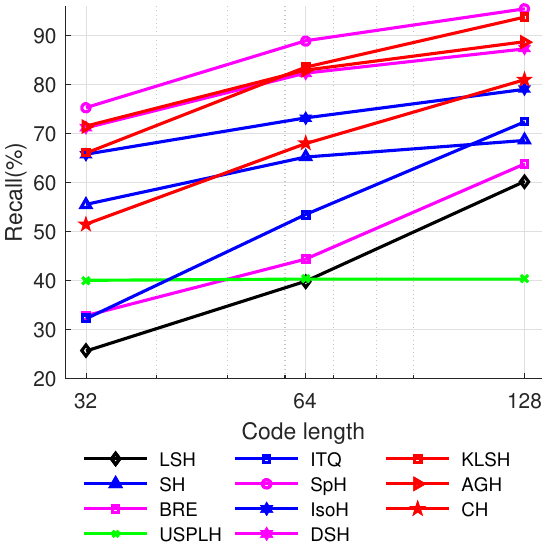}\includegraphics[width=\SmallDoubleFigureWidth]{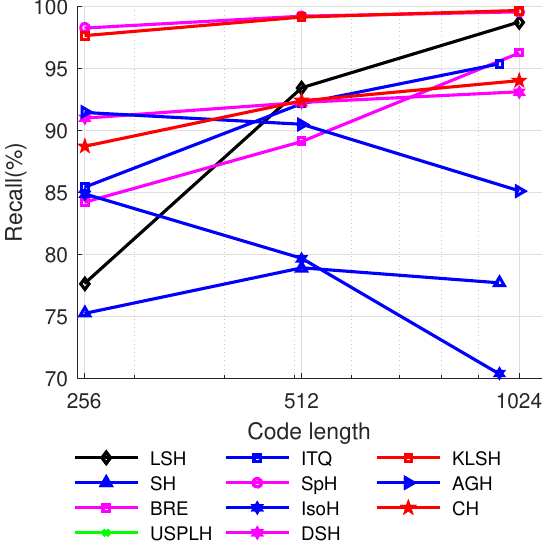}
	\caption{The {\bf recall - codelength} curves of various hashing algorithms when locate 50000 samples on GIST1M}
	\label{PvsCodelengthGIST}
\end{figure*}
	
	\item {\bf SpH} is a short name for Spherical Hashing \cite{HeoLHCY12CVPR}. SpH uses a hyperspherebased
	hash function to map data points into binary codes.
	\item {\bf IsoH} is a short name for Isotropic Hashing \cite{KongL12NIPS}. IsoH learns the
	projection functions with isotropic variances for PCA projected	data. The main motivation of IsoH is that PCA directions with different variance should not be equally treated (one bit for one direction).
	\item {\bf CH} is a short name for Compressed Hashing \cite{LinJCYL13CVPR}. CH first learns a landmark (anchor) based sparse representation \cite{ChenC11AAAI} then followed by LSH.
	\item {\bf DSH} is a short name for Density Sensitive Hashing \cite{JinLLC14TCYB}. DSH finds the projections which aware the density distribution of the data. 
\end{itemize}
For SH, ITQ, and IsoH, the learning process will involve the computation of eigenvectors of the covariance matrix. Thus, these three algorithms can only learn 128-bits code on SIFT1M and 960-bits code on GIST1M. All the other eight algorithms learn 1,024-bits code on both datasets.

For KLSH, AGH and CH, we need to pick a certain number of anchor (landmark) points. We use the same 1,500 anchor points for all these algorithms, and the anchor points are generated by using kmeans with five iterations. Also, the number of nearest anchors is set to 50 for AGH and CH algorithms. Please see \cite{LiuWKC11ICML} for details.

All the hashing algorithms are implemented in Matlab\footnote{Almost all the core parts of the Matlab code of the hashing algorithms are written by the original authors of the papers.}, and we use these hashing functions to learn the binary codes for both base vectors and query vectors. The binary codes can then be fed into Algorithm \ref{alg:Qhamming} for search. We use the same 1,000 clusters generated by kmeans for all the hashing algorithms. The Matlab functions are run on an i7-5930K CPU with 128G memory, and the Algorithm \ref{alg:Qhamming} is implemented in c++ and run on an i7-4790K CPU and 32G memory.  For the sake of reproducibility, both the Matlab codes and c++ codes are released on GitHub\footnote{https://github.com/ZJULearning/hashingSearch}.

Besides all the hashing algorithms, we also report the performance of the brute-force search as a baseline.
\begin{itemize}
	\item {\bf brute-force}. The performance of brute-force search is reported to show the advantages of using hashing methods. To get different recall, we simply perform  brute-force search on different percentage of the query number. For example, the brute-force search time on 90\% queries of the origin query set stands for the brute-force search time of 90\% recall. 
\end{itemize}

\begin{figure*}
	\centering
	\subfigure[]{\includegraphics[width=\DoubleFigureWidth]{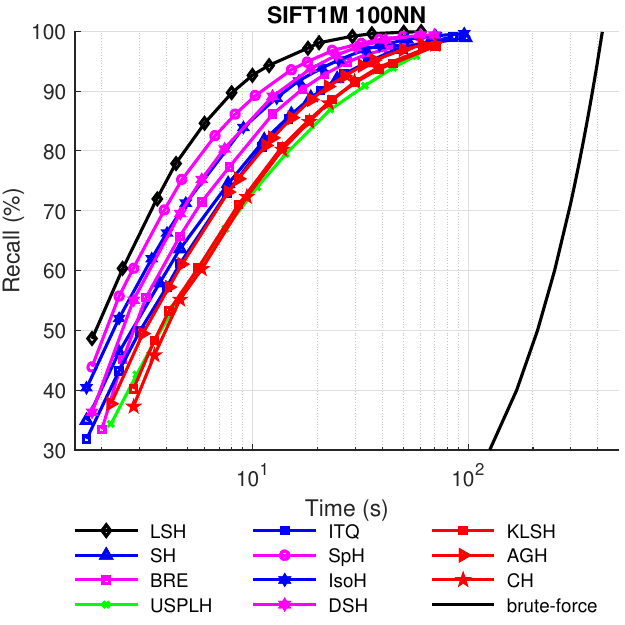}}
	\subfigure[]{\includegraphics[width=\DoubleFigureWidth]{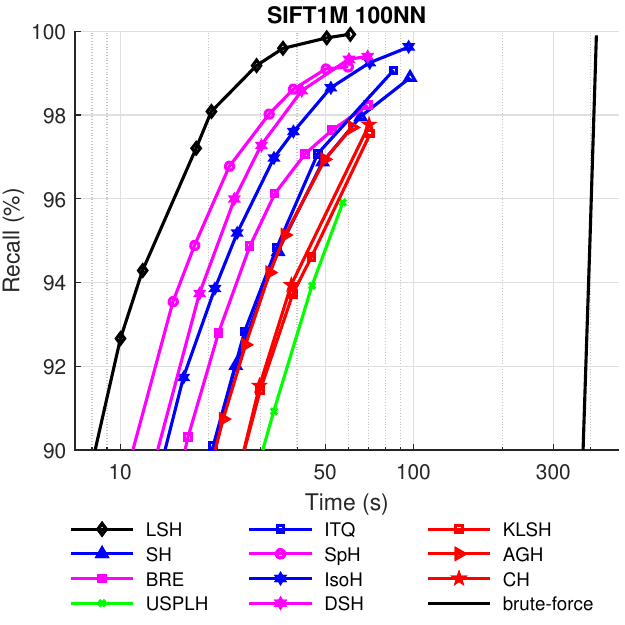}}
	\caption{The {\bf recall-time} curves of various hashing algorithms using the grouped hamming ranking approach on SIFT1M. }
	\label{PvsTimeBestSIFT}
\end{figure*}

\begin{figure*}
	\centering
	\subfigure[]{\includegraphics[width=\DoubleFigureWidth]{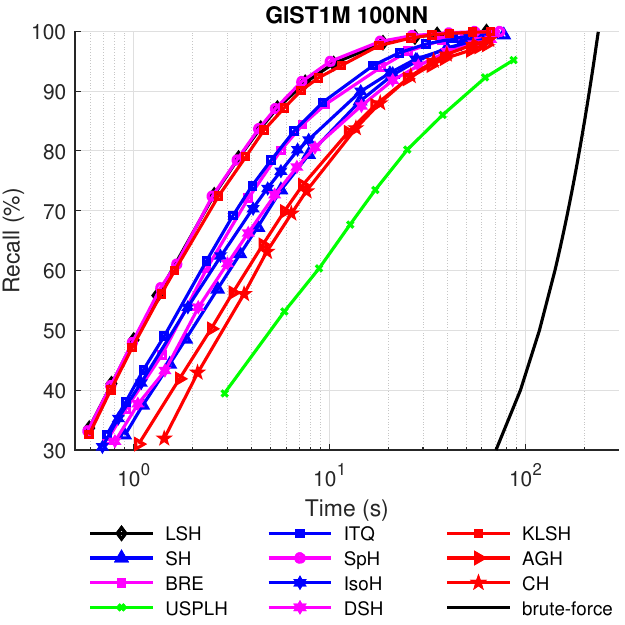}}
	\subfigure[]{\includegraphics[width=\DoubleFigureWidth]{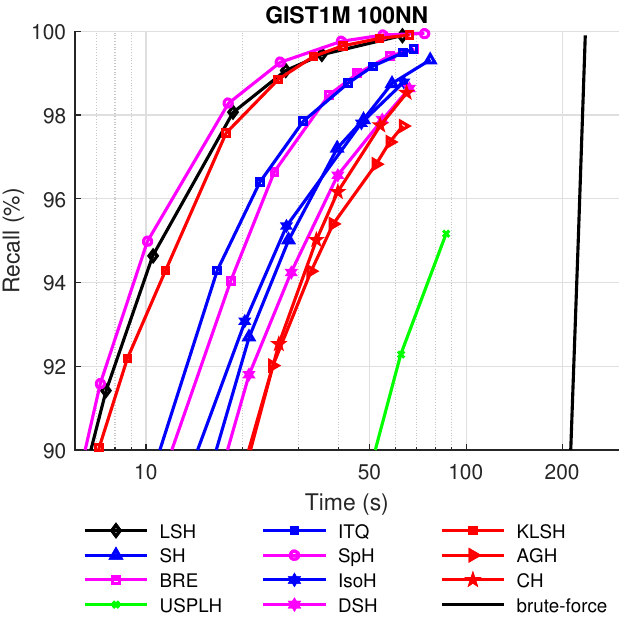}}
	\caption{The {\bf recall-time} curves of various hashing algorithms using the grouped hamming ranking approach on GIST1M. }
	\label{PvsTimeBestGIST}
\end{figure*}

\begin{figure*}
	\centering
	\subfigure[]{\includegraphics[width=\DoubleFigureWidth]{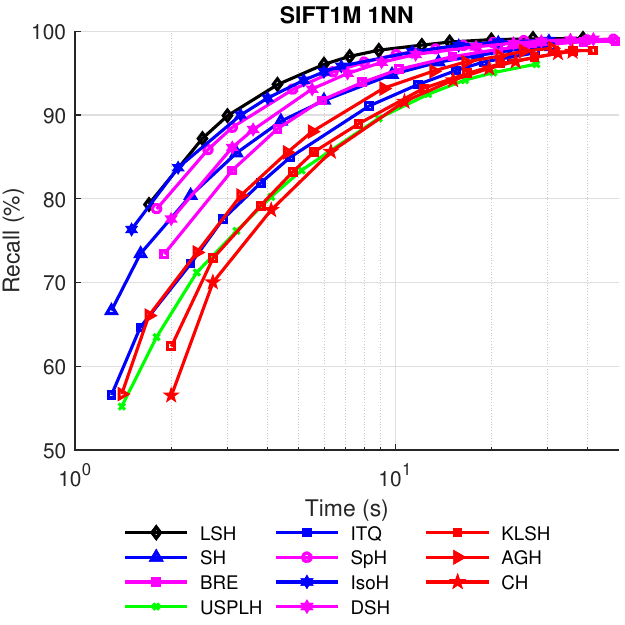}}
	\subfigure[]{\includegraphics[width=\DoubleFigureWidth]{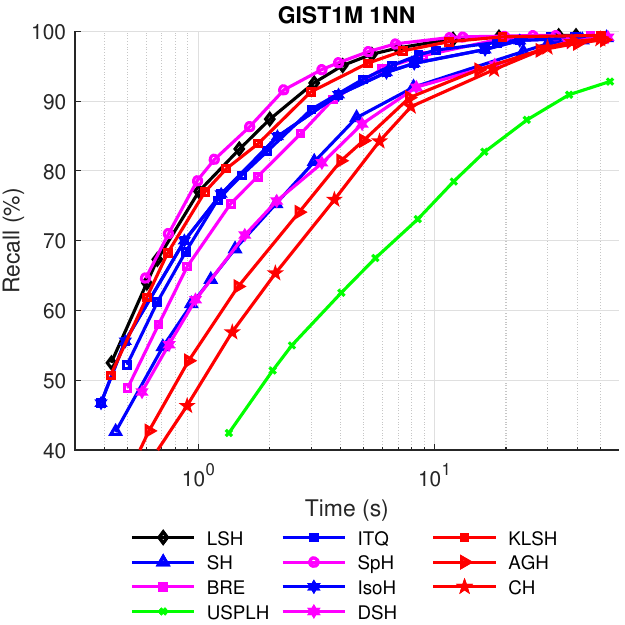}}
	\caption{The {\bf recall-time} curves of various hashing algorithms using the grouped hamming ranking approach with $k=1$. }
	\label{PvsTimeBestK1}
\end{figure*}

\begin{figure*}
	\centering
	\subfigure[]{\includegraphics[width=\DoubleFigureWidth]{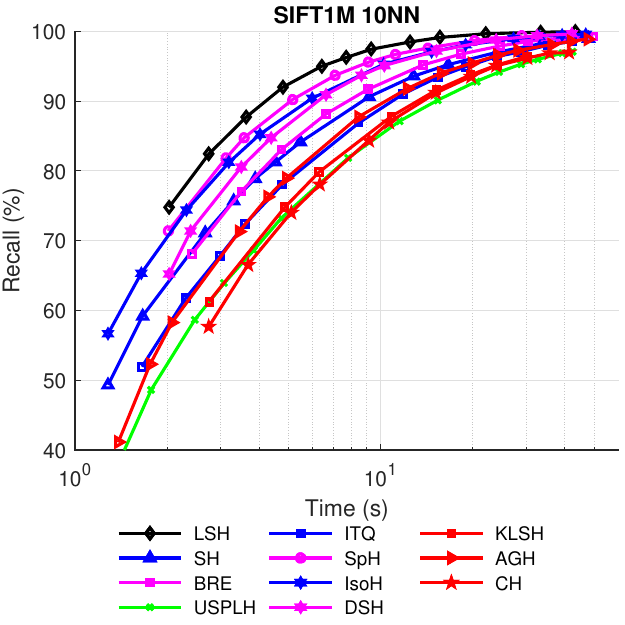}}
	\subfigure[]{\includegraphics[width=\DoubleFigureWidth]{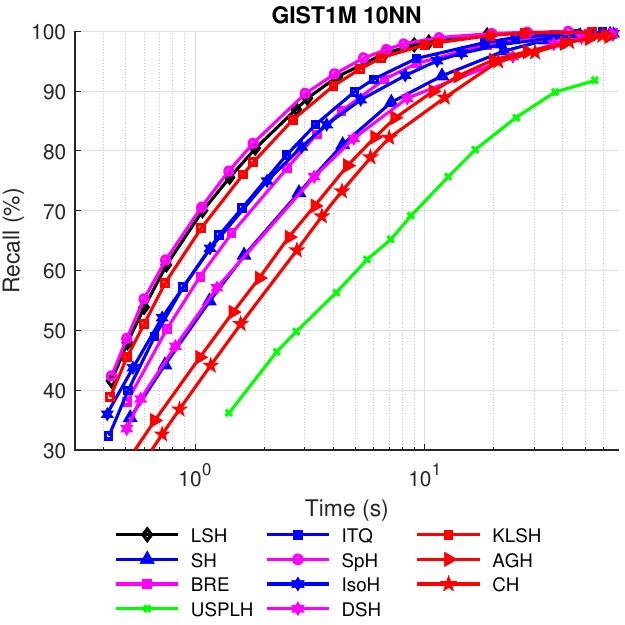}}
	\caption{The {\bf recall-time} curves of various hashing algorithms using the grouped hamming ranking approach with $k=10$. }
	\label{PvsTimeBestK10}
\end{figure*}

\subsection{Results}

Since we use Matlab functions to learn the binary codes and c++ algorithm to search with a hash index, the coding time and later time (locating time and scanning time) cannot be added together, and we reported them separately. 

Table \ref{indexingtime} and \ref{codingtime} report the indexing (hash learning) time and coding (testing) time of various hashing algorithms. The indexing (hash learning) stage is performed off-line thus the indexing time is not very crucial. The coding time of all the hashing algorithms is short. Therefore the coding time can be ignored compared with locating time and scanning time (see Figure \ref{PvsTimeBestSIFT} and \ref{PvsTimeBestGIST}) when we consider the total search time.

Figure \ref{PvsCodelengthSIFT} and \ref{PvsCodelengthGIST} show how the recall changes as the code length increases of various hashing algorithms when the number of the located samples are fixed at 10,000 on SIFT1M and 50,000 on GIST1M. This figure illustrates several interesting points:
\begin{itemize}
	\item When the code length is 32, all of the other algorithms outperform LSH on both two datasets. When the code length is 64, eight algorithms outperform LSH on SIFT1M, and nine algorithms outperform LSH on GIST1M. However, when the code length exceeds 512, LSH performs the best on SIFT1M and the 3rd best on GIST1M. This result is consistent with the finding in \cite{JolyB11CVPR} that "many data-dependent hashing algorithms have the improvements over LSH, but improvements occur only for relatively small hash code sizes up to 64 or 128 bits". This result is also consistent with the "outperform LSH" claim in all these hashing papers since they never report the performance using longer codes.
	\item If the locating time is ignored, the performance of most of the algorithms increases as the code length increases. However, on SIFT1M, the recall of USPLH decreases as the code length exceeds 256 and the recall of AGH decreases as the code length exceeds 512. On GIST1M, the recall of USPLH almost fixed as the code length increases and the recalls of AGH, SH and IsoH decrease as the code length exceeds 512. This result suggests the four hashing algorithms, USPLH, AGH, SH, and IsoH may not be able to learn very long discriminative codes.
\end{itemize}

We cannot judge which algorithm is better based merely on Figure \ref{PvsCodelengthSIFT} and \ref{PvsCodelengthGIST}, since the locating time is ignored. To pick the best hashing algorithm, we have to use the hashing code as the index and use {\bf time - recall} curve to see the ANNS performance.

For all the hashing algorithms, we use the Algorithm \ref{alg:Qhamming} (the grouped hamming ranking approach) for search. All the algorithms use the same 1,000 clusters generated by kmeans. Based on the Figure \ref{PvsCodelengthSIFT} and \ref{PvsCodelengthGIST}, we can pick the optimal code length of each hash algorithm. On SIFT1M, SH, ITQ and IsoH use 128 bits code, AGH and USPLH use 256 bits code, and all the other hashing algorithms use 1,024 bits code. On GIST1M, AGH and IsoH use 256 bits code, ITQ uses 960 bits code, USPLH uses 64 bits code, and all the other hashing algorithms use 1,024 bits code.

\begin{figure*}
	\centering
	\subfigure[]{\includegraphics[width=\DoubleFigureWidth]{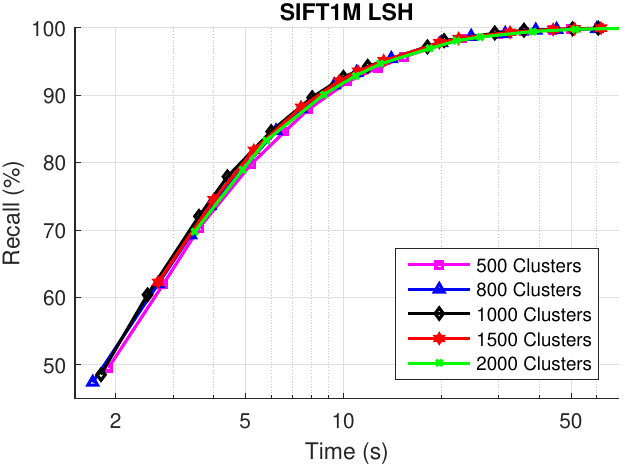}}
	\subfigure[]{\includegraphics[width=\DoubleFigureWidth]{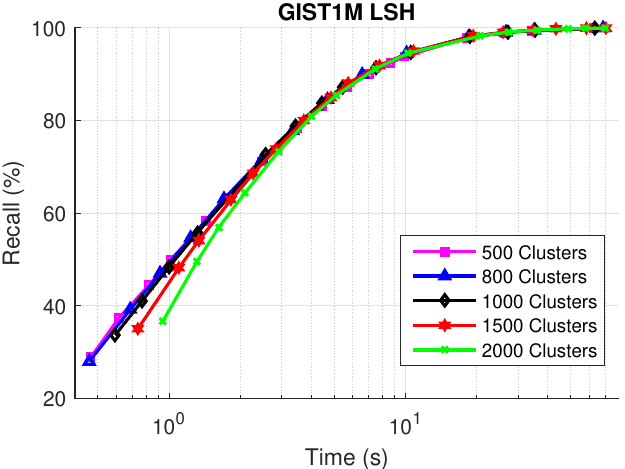}}
	\caption{The performance of LSH using the grouped hamming ranking approach with various cluster number. }
	\label{PvsTimeCluster}
\end{figure*}

\begin{figure*}
	\centering
	\subfigure[]{\includegraphics[width=\DoubleFigureWidth]{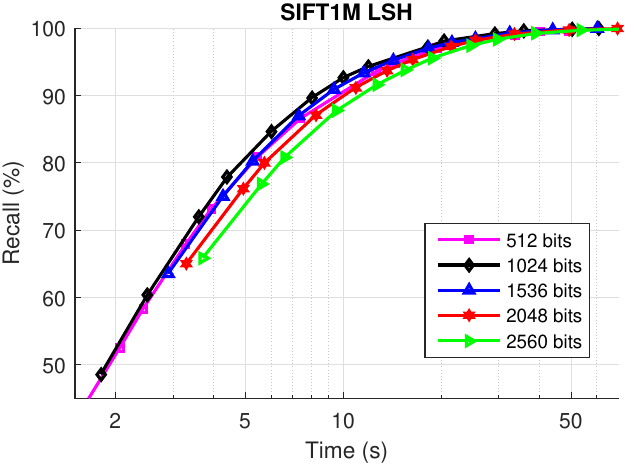}}
	\subfigure[]{\includegraphics[width=\DoubleFigureWidth]{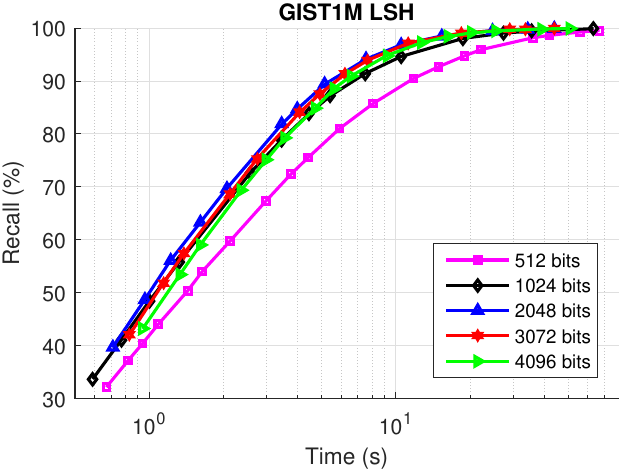}}
	\caption{The performance of LSH using the grouped hamming ranking approach with various code length. }
	\label{PvsTimeCodeLength}
\end{figure*}

Figure \ref{PvsTimeBestSIFT} and \ref{PvsTimeBestGIST} show the {\bf recall-time} curves of various hashing algorithms using the grouped hamming ranking approach on SIFT1M and GIST1M respectively. A number of interesting points can be found:
On SIFT1M, LSH performs significantly better than all the other hashing algorithms. On GIST1M, LSH is the second best-performed algorithm. It is slightly worse than SpH. Considering the extreme simplicity of LSH, we can conclude that LSH is the best choice among all the eleven compared hashing algorithms. This finding is a contradiction to the conclusions in most of the previous hashing papers. 

In the previous and the remaining experiments, we require all the algorithms to return 100 answers, \ie, fix $K=100$. To show this setting is reasonable, we plot the {\bf recall-time} curves of various hashing algorithms with $K=1$ and $K=10$ in the Figure \ref{PvsTimeBestK1} and \ref{PvsTimeBestK10}. We can see the same trend as $K=100$. As $K=1, 10, 100$, LSH is always the best performed algorithm on SIFT1M and the second best-performed algorithm on GIST1M.

\subsection{Parameters of The Grouped Hamming Ranking}

There are two essential parameters in the proposed grouped hamming ranking approach: the number of groups $G$ and the binary code length $l$. In this Section, we will discuss how these two parameters affect the performance of the grouped hamming ranking approach.

Figure \ref{PvsTimeCluster} shows the performance of LSH on SIFT1M and GIST1M with various number of clusters (500, 800, 1,000, 1,500 and 2,000) using the grouped hamming ranking. We can see the search performance is very stable. With 1M points, 1,000 clusters means the average number of points in each cluster is 1,000. This is a reasonable choice. 

The binary code length is another important parameter. In the previous experiment, we choose this parameter for various hashing algorithms based on the {\em recall-codelength} curve (See Figure \ref{PvsCodelengthSIFT} and \ref{PvsCodelengthGIST}). And we set the maximum length to be 1,024. It turns out that most of the compared hashing algorithms choose this parameter as 1,024. How about we further increase the code length? Figure \ref{PvsTimeCodeLength} shows the performance of LSH on SIFT1M and GIST1M with various code length. On SIFT1M, LSH with grouped hamming ranking reaches the best search performance as the code length is 1,024. The search performance slowly decreases as the code length further increases.  On GIST1M, the optimal code length becomes 2,048.

\begin{figure*}
	\centering
	\subfigure[SIFT1M]{\includegraphics[width=\DoubleFigureWidth]{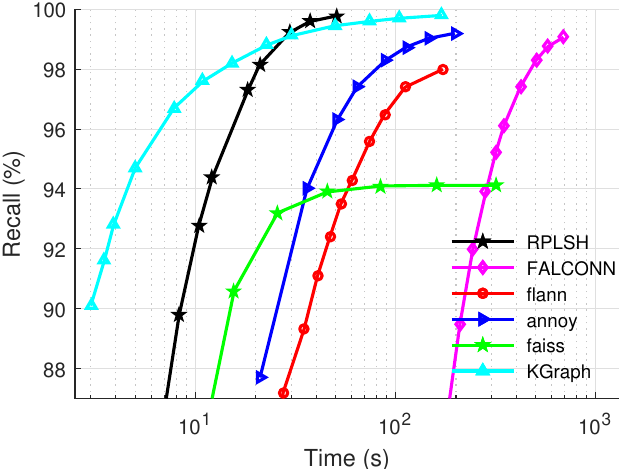}}
	\subfigure[GIST1M]{\includegraphics[width=\DoubleFigureWidth]{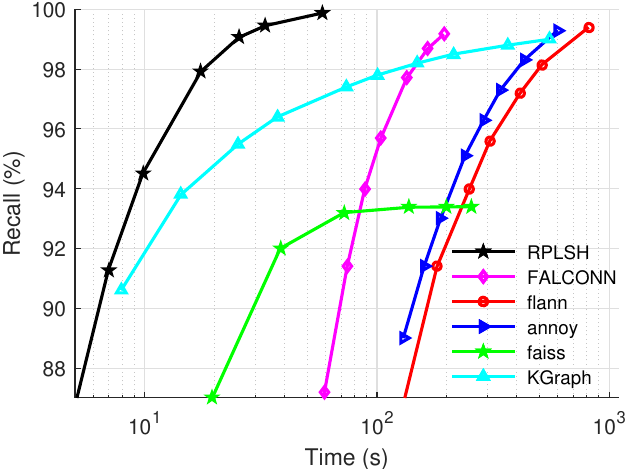}}
	\caption{The {\bf recall-time} curves of six popular ANNS methods on SIFT1M and GIST1M.}
	\label{ANNS}
\end{figure*}

\section{Comparison with other popular ANNS methods}

Given the surprising performance of random-projection-based LSH, we want to compare it with some public popular ANNS software. To be fair, we convert our Matlab code for random-projection-based LSH to c++ code and make a complete search software (by c++). We named our algorithm RPLSH. The codes are also released on GitHub\footnote{https://github.com/ZJULearning/RPLSH}.

\subsection{Compared Algorithms}

We pick five popular ANNS algorithms for comparison which cover various types such as tree-based, hashing-based, quantization-based and graph-based approaches. They are:
\begin{enumerate}
	\item \textbf{FALCONN}\footnote{https://github.com/FALCONN-LIB/FALCONN} is a well-known ANNS library based on multi-probe locality sensitive hashing. It implements the hash bucket search with multiple tables approach.
	\item \textbf{flann} is a well-known ANNS library based on trees \cite{Muja2014Scalable}. It integrates Randomized KD-tree \cite{Silpaanan2008Optimised} and Kmeans tree \cite{Fukunaga1975A}. In our experiment, we use the randomized KD-tree and set the number of the trees as 32 on SIFT1M and 64 on GIST1M. The code on GitHub can be found at\footnote{https://github.com/mariusmuja/flann}.
	\item \textbf{annoy} is based on a binary search forest index. We use their code on GitHub for comparison\footnote{https://github.com/spotify/annoy}.
	\item \textbf{faiss} is recently released\footnote{https://github.com/facebookresearch/faiss} by Facebook. It contains well implemented code for state-of-the-art product quantization \cite{JegouDS11PAMI} based methods both on CPU and GPU. The CPU version is used for a fair comparison.
	\item \textbf{KGraph} has open source code GitHub at\footnote{https://github.com/aaalgo/kgraph}, which is based on an approximate $k$NN Graph. 
\end{enumerate}

The experiments are carried out on a machine with an i7-4790k CPU and 32G memory. The performance on the high recall region for an ANNS algorithm will be more crucial in real applications. Thus we sample one percentage points from each base
set as its corresponding validation set and tune all the algorithms on the validation sets to get their best-performing indices at the high recall region. For RPLSH, we use 1,024-bits code and 1,000 clusters. For all the search experiments, we only evaluate the algorithms on a single thread.

\subsection{Results}

The {\bf recall-time} curves of the six compared ANNS methods are shown in Figure \ref{ANNS}. A number of interesting conclusions can be made:
\begin{itemize}
	\item On both two datasets, RPLSH performs at least the second best among all the six compared algorithms when the recall is higher than 87\%. On SIFT1M, If the recall is higher than 99\% on SIFT1M and higher than 90\% on GIST1M,  RPLSH performs the best. 
	\item Both the RPLSH and the FALCONN are LSH based algorithms. RPLSH significantly outperforms FALCONN on both two datasets. This probably dues to the reason that RPLSH uses the grouped hamming ranking approach while FALCONN uses the hash buckets search with multiple tables approach.
	\item flann and annoy are tree-based methods. Compare the performance of two hashing-based methods with that of two tree-based methods, and we can find that tree-based methods are more suitable for low dimensional data.
	\item faiss is a product quantization (PQ) \cite{JegouDS11PAMI} based method, and it uses binary codes to approximate the original features. Thus it is impossible for faiss to achieve a very high recall. The advantage of this approximation is that we do not need the original vectors which is a huge saving of the memory usage \cite{JegouDS11PAMI}. 
	\item KGraph is a graph-based algorithm, and it is KGraph's author who says "LSH is so hopelessly slow and/or inaccurate"\footnote{http://www.kgraph.org/}. However, RPLSH is significantly better than KGraph on GIST1M when the recall is higher than 90\%.
\end{itemize}

\section{Conclusions}

We carefully studied the problem of using hashing algorithms for ANNS. We introduce a novel approach (grouped hamming ranking) to search with a hash index. With this new approach, random-projection-based LSH is superior to many other popular hashing algorithms, and this algorithm is extremely simple. All the codes used in the paper are released on GitHub, which can be used as a testing platform for a fair comparison between various hashing methods.


\begin{thebibliography}{10}\itemsep=-1pt
	
	\bibitem{AryaM93Approximate}
	S.~Arya and D.~M. Mount.
	\newblock Approximate nearest neighbor queries in fixed dimensions.
	\newblock In {\em Proceedings of the Fourth Annual {ACM/SIGACT-SIAM} Symposium
		on Discrete Algorithms}, 1993.
	
	\bibitem{Belkin01Eigenmap}
	M.~Belkin and P.~Niyogi.
	\newblock Laplacian eigenmaps and spectral techniques for embedding and
	clustering.
	\newblock In {\em Advances in Neural Information Processing Systems 14}, 2001.
	
	\bibitem{Ben2016Fanng}
	H.~Ben and D.~Tom.
	\newblock {FANNG}: Fast approximate nearest neighbour graphs.
	\newblock In {\em Proceedings of the 2016 IEEE Conference on Computer Vision
		and Pattern Recognition}, pages 5713--5722, 2016.
	
	\bibitem{Bentley1975Multidimensional}
	J.~L. Bentley.
	\newblock Multidimensional binary search trees used for associative searching.
	\newblock {\em Communications of the ACM}, 18(9):509--517, 1975.
	
	\bibitem{Charikar2002LSH}
	M.~S. Charikar.
	\newblock Similarity estimation techniques from rounding algorithms.
	\newblock In {\em Proceedings of the Thiry-fourth Annual ACM Symposium on
		Theory of Computing}, pages 380--388, 2002.
	
	\bibitem{ChenC11AAAI}
	X.~Chen and D.~Cai.
	\newblock Large scale spectral clustering with landmark-based representation.
	\newblock In {\em Proceedings of the Twenty-Fifth {AAAI} Conference on
		Artificial Intelligence}, 2011.
	
	\bibitem{fu2017fast}
	C.~Fu, C.~Xiang, C.~Wang, and D.~Cai.
	\newblock Fast approximate nearest neighbor search with the navigating
	spreading-out graphs.
	\newblock {\em {PVLDB}}, 12(5):461 -- 474, 2019.
	
	\bibitem{Fukunaga1975A}
	K.~Fukunaga and P.~M. Narendra.
	\newblock A branch and bound algorithm for computing k-nearest neighbors.
	\newblock {\em IEEE Transactions on Computers}, 100(7):750--753, 1975.
	
	\bibitem{GeHK013CVPR}
	T.~Ge, K.~He, Q.~Ke, and J.~Sun.
	\newblock Optimized product quantization for approximate nearest neighbor
	search.
	\newblock In {\em 2013 {IEEE} Conference on Computer Vision and Pattern
		Recognition}, 2013.
	
	\bibitem{GiVLDBIM1999VLDB}
	A.~Gionis, P.~Indyk, and R.~Motwani.
	\newblock Similarity search in high dimensions via hashing.
	\newblock In {\em Proceedings of the 25th International Conference on Very
		Large Data Bases}, 1999.
	
	\bibitem{GongL11CVPR}
	Y.~Gong and S.~Lazebnik.
	\newblock Iterative quantization: {A} procrustean approach to learning binary
	codes.
	\newblock In {\em The 24th {IEEE} Conference on Computer Vision and Pattern
		Recognition}, 2011.
	
	\bibitem{HeLC10KDD}
	J.~He, W.~Liu, and S.~Chang.
	\newblock Scalable similarity search with optimized kernel hashing.
	\newblock In {\em Proceedings of the 16th {ACM} International Conference on
		Knowledge Discovery and Data Mining}, 2010.
	
	\bibitem{HeWS13CVPR}
	K.~He, F.~Wen, and J.~Sun.
	\newblock K-means hashing: An affinity-preserving quantization method for
	learning binary compact codes.
	\newblock In {\em 2013 {IEEE} Conference on Computer Vision and Pattern
		Recognition}, 2013.
	
	\bibitem{HeoLHCY12CVPR}
	J.~Heo, Y.~Lee, J.~He, S.~Chang, and S.~Yoon.
	\newblock Spherical hashing.
	\newblock In {\em 2012 {IEEE} Conference on Computer Vision and Pattern
		Recognition}, 2012.
	
	\bibitem{IndykM98STOC}
	P.~Indyk and R.~Motwani.
	\newblock Approximate nearest neighbors: Towards removing the curse of
	dimensionality.
	\newblock In {\em Proceedings of the Thirtieth Annual {ACM} Symposium on the
		Theory of Computing}, 1998.
	
	\bibitem{JegouDS11PAMI}
	H.~J{\'{e}}gou, M.~Douze, and C.~Schmid.
	\newblock Product quantization for nearest neighbor search.
	\newblock {\em {IEEE} Trans. Pattern Anal. Mach. Intell.}, 33(1):117--128,
	2011.
	
	\bibitem{JinHLZLCL13ICCV}
	Z.~Jin, Y.~Hu, Y.~Lin, D.~Zhang, S.~Lin, D.~Cai, and X.~Li.
	\newblock Complementary projection hashing.
	\newblock In {\em {IEEE} International Conference on Computer Vision}, 2013.
	
	\bibitem{JinLLC14TCYB}
	Z.~Jin, C.~Li, Y.~Lin, and D.~Cai.
	\newblock Density sensitive hashing.
	\newblock {\em {IEEE} Trans. Cybernetics}, 44(8):1362--1371, 2014.
	
	\bibitem{Jin2014Fast}
	Z.~Jin, D.~Zhang, Y.~Hu, S.~Lin, D.~Cai, and X.~He.
	\newblock Fast and accurate hashing via iterative nearest neighbors expansion.
	\newblock {\em IEEE transactions on cybernetics}, 44(11):2167--2177, 2014.
	
	\bibitem{JolyB11CVPR}
	A.~Joly and O.~Buisson.
	\newblock Random maximum margin hashing.
	\newblock In {\em The 24th {IEEE} Conference on Computer Vision and Pattern
		Recognition}, 2011.
	
	\bibitem{KalantidisA14CVPR}
	Y.~Kalantidis and Y.~S. Avrithis.
	\newblock Locally optimized product quantization for approximate nearest
	neighbor search.
	\newblock In {\em 2014 {IEEE} Conference on Computer Vision and Pattern
		Recognition}, 2014.
	
	\bibitem{Kleinberg97Algorithms}
	J.~M. Kleinberg.
	\newblock Two algorithms for nearest-neighbor search in high dimensions.
	\newblock In {\em Proceedings of the Twenty-Ninth Annual {ACM} Symposium on the
		Theory of Computing}, 1997.
	
	\bibitem{KongL12NIPS}
	W.~Kong and W.~Li.
	\newblock Isotropic hashing.
	\newblock In {\em Advances in Neural Information Processing Systems 25}, 2012.
	
	\bibitem{KulisD09NIPS}
	B.~Kulis and T.~Darrell.
	\newblock Learning to hash with binary reconstructive embeddings.
	\newblock In {\em Advances in Neural Information Processing Systems 22,
		{NIPS}}, 2009.
	
	\bibitem{KulisG09ICCV}
	B.~Kulis and K.~Grauman.
	\newblock Kernelized locality-sensitive hashing for scalable image search.
	\newblock In {\em {IEEE} 12th International Conference on Computer Vision},
	2009.
	
	\bibitem{li2016approximate}
	W.~Li, Y.~Zhang, Y.~Sun, W.~Wang, W.~Zhang, and X.~Lin.
	\newblock Approximate nearest neighbor search on high dimensional
	data---experiments, analyses, and improvement (v1. 0).
	\newblock {\em arXiv:1610.02455}, 2016.
	
	\bibitem{LinJCYL13CVPR}
	Y.~Lin, R.~Jin, D.~Cai, S.~Yan, and X.~Li.
	\newblock Compressed hashing.
	\newblock In {\em 2013 {IEEE} Conference on Computer Vision and Pattern
		Recognition}, 2013.
	
	\bibitem{Liu10AnchorGraph}
	W.~Liu, J.~He, and S.-F. Chang.
	\newblock Large graph construction for scalable semi-supervised learning.
	\newblock In {\em Proceedings of the 27th International Conference on Machine
		Learning}, 2010.
	
	\bibitem{LiuMKC14NIPS}
	W.~Liu, C.~Mu, S.~Kumar, and S.~Chang.
	\newblock Discrete graph hashing.
	\newblock In {\em Advances in Neural Information Processing Systems 27}, 2014.
	
	\bibitem{LiuWJJC12CVPR}
	W.~Liu, J.~Wang, R.~Ji, Y.~Jiang, and S.~Chang.
	\newblock Supervised hashing with kernels.
	\newblock In {\em 2012 {IEEE} Conference on Computer Vision and Pattern
		Recognition}, 2012.
	
	\bibitem{LiuWKC11ICML}
	W.~Liu, J.~Wang, S.~Kumar, and S.~Chang.
	\newblock Hashing with graphs.
	\newblock In {\em Proceedings of the 28th International Conference on Machine
		Learning}, 2011.
	
	\bibitem{Makhoul2000Performance}
	J.~Makhoul, F.~Kubala, R.~Schwartz, and R.~Weischedel.
	\newblock Performance measures for information extraction.
	\newblock In {\em Proceedings of DARPA Broadcast News Workshop}, 2000.
	
	\bibitem{MalkovNMSLIB2016}
	Y.~A. {Malkov} and D.~A. {Yashunin}.
	\newblock Efficient and robust approximate nearest neighbor search using
	hierarchical navigable small world graphs.
	\newblock {\em arXiv:1603.09320}, 2016.
	
	\bibitem{Muja2014Scalable}
	M.~Muja and D.~G. Lowe.
	\newblock Scalable nearest neighbor algorithms for high dimensional data.
	\newblock {\em IEEE Transactions on Pattern Analysis and Machine Intelligence},
	36(11):2227--2240, 2014.
	
	\bibitem{NorouziF11ICML}
	M.~Norouzi and D.~J. Fleet.
	\newblock Minimal loss hashing for compact binary codes.
	\newblock In {\em Proceedings of the 28th International Conference on Machine
		Learning,{ICML} 2011}, 2011.
	
	\bibitem{Silpaanan2008Optimised}
	C.~Silpa-Anan and R.~Hartley.
	\newblock Optimised kd-trees for fast image descriptor matching.
	\newblock In {\em Proceedings of the 2008 IEEE Conference on Computer Vision
		and Pattern Recognition}, 2008.
	
	\bibitem{WangKC10ICML}
	J.~Wang, S.~Kumar, and S.~Chang.
	\newblock Sequential projection learning for hashing with compact codes.
	\newblock In {\em Proceedings of the 27th International Conference on Machine
		Learning}, 2010.
	
	\bibitem{Hacker}
	H.~S. Warren.
	\newblock {\em Hacker's Delight, Chapter 5}.
	\newblock Addison-Wesley Professional, 2012.
	
	\bibitem{Weber1998AQA}
	R.~Weber, H.-J. Schek, and S.~Blott.
	\newblock A quantitative analysis and performance study for similarity-search
	methods in high-dimensional spaces.
	\newblock In {\em VLDB}, 1998.
	
	\bibitem{WeissFT12}
	Y.~Weiss, R.~Fergus, and A.~Torralba.
	\newblock Multidimensional spectral hashing.
	\newblock In {\em 12th European Conference on Computer Vision}, 2012.
	
	\bibitem{WeissTF08NIPS}
	Y.~Weiss, A.~Torralba, and R.~Fergus.
	\newblock Spectral hashing.
	\newblock In {\em Advances in Neural Information Processing Systems 21,
		{NIPS}}, 2008.
	
	\bibitem{XuBLCHC13IJCAI}
	B.~Xu, J.~Bu, Y.~Lin, C.~Chen, X.~He, and D.~Cai.
	\newblock Harmonious hashing.
	\newblock In {\em Proceedings of the 23rd International Joint Conference on
		Artificial Intelligence}, 2013.
	
	\bibitem{XuWLZLY11ICCV}
	H.~Xu, J.~Wang, Z.~Li, G.~Zeng, S.~Li, and N.~Yu.
	\newblock Complementary hashing for approximate nearest neighbor search.
	\newblock In {\em {IEEE} International Conference on Computer Vision}, 2011.
	
	\bibitem{ZhangWCL10SIGIR}
	D.~Zhang, J.~Wang, D.~Cai, and J.~Lu.
	\newblock Self-taught hashing for fast similarity search.
	\newblock In {\em Proceeding of the 33rd International {ACM} {SIGIR} Conference
		on Research and Development in Information Retrieval}, 2010.
	
\end{thebibliography}
\end{document}